\documentclass[10pt,twocolumn,letterpaper]{article}

\usepackage{cvpr}
\usepackage{times}
\usepackage{epsfig}
\usepackage{graphicx}
\usepackage{amsmath}
\usepackage{amssymb}
\usepackage{gensymb}

\usepackage{adjustbox}
\newcommand{\widthscalefive}{0.10}
\newcommand{\widthscalesix}{0.12}

\usepackage{pifont}% http://ctan.org/pkg/pifont
\newcommand{\xmark}{\ding{55}}%
\cvprfinalcopy % *** Uncomment this line for the final submission

\def\cvprPaperID{7376} % *** Enter the CVPR Paper ID here

\ifcvprfinal\fi
\begin{document}

%%%%%%%%% TITLE
\title{Spatially-Attentive Patch-Hierarchical Network\\ for Adaptive Motion Deblurring}

\author{Maitreya Suin$^*$ \qquad Kuldeep Purohit$\thanks{Equal contribution.}$ \qquad A. N. Rajagopalan \\
Indian Institute of Technology Madras, India\\
%{\tt\small \{chi.zhang,f.gao,baoxiongjia,yixin.zhu\}@ucla.edu, sczhu@stat.ucla.edu}
{\tt\small maitreyasuin21@gmail.com, kuldeeppurohit3@gmail.com, raju@ee.iitm.ac.in}
% For a paper whose authors are all at the same institution,
% omit the following lines up until the closing ``}''.
% Additional authors and addresses can be added with ``\and'',
% just like the second author.
% To save space, use either the email address or home page, not both
} 

\maketitle
%\thispagestyle{empty}

%%%%%%%%% ABSTRACT
\begin{abstract}
This paper tackles the problem of motion deblurring of dynamic scenes. Although end-to-end fully convolutional designs have recently advanced the state-of-the-art in non-uniform motion deblurring, their performance-complexity trade-off is still sub-optimal. Existing approaches achieve a large receptive field by increasing the number of generic convolution layers and kernel-size, but this comes at the expense of of the increase in model size and inference speed. In this work, we propose an efficient pixel adaptive and feature attentive design for handling large blur variations across different spatial locations and process each test image adaptively. We also propose an effective content-aware global-local filtering module that significantly improves performance by considering not only global dependencies but also by dynamically exploiting neighboring pixel information. We use a patch-hierarchical attentive architecture composed of the above module that implicitly discovers the spatial variations in the blur present in the input image and in turn, performs local and global modulation of intermediate features. Extensive qualitative and quantitative comparisons with prior art on deblurring benchmarks demonstrate that our design offers significant improvements over the state-of-the-art in accuracy as well as speed.

\end{abstract}

%%%%%%%%% BODY TEXT
\section{Introduction}
Motion-blurred images form due to relative motion during sensor exposure and are favored by photographers and artists in many cases for aesthetic purpose, but seldom by computer vision researchers, as many standard vision tools including detectors, trackers, and feature extractors struggle to deal with blur. Blind motion deblurring is an ill-posed problem that aims to recover a sharp image from a given image degraded due to motion-induced smearing of texture and high-frequency details. Due to its diverse applications in surveillance, remote sensing, and cameras mounted on hand-held and vehicle-mounted cameras, deblurring has gathered substantial attention from computer vision and image processing communities in the past two decades. 

\begin{figure}[t] \label{fig:intro}
\begin{center}
   \includegraphics[width=1\linewidth]{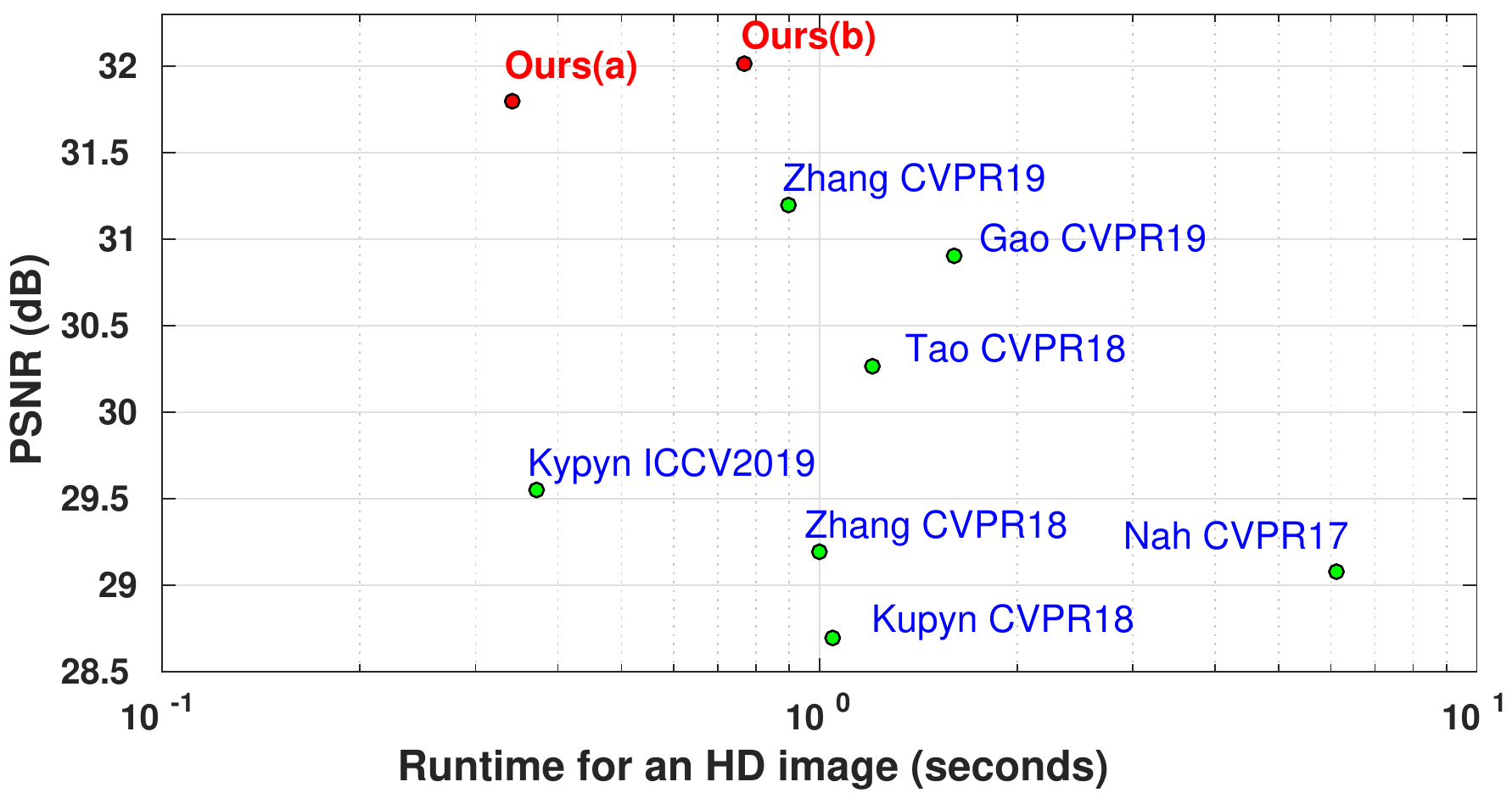}
\end{center}
   \caption{Comparison of different methods in terms of accuracy and inference time. Our approach outperforms all previous methods.}
\label{fig:long}
\label{fig:onecol}
\end{figure}

Majority of traditional deblurring approaches are based on variational model, whose key component is the regularization term. The restoration quality depends on the selection of the prior, its weight, as well as tuning of other parameters involving highly non-convex optimization setups\cite{nimisha2017blur}. Non-uniform blind deblurring for general dynamic scenes is a challenging computer vision problem as blurs arise from various sources including moving objects, camera shake and depth variations, causing different pixels to capture different motion trajectories. Such hand-crafted priors struggle while generalizing across different types of real-world examples, where blur is far more complex than modeled \cite{gong2017motion}. 

% Regarding motion trajectory, various conventional and learning based approaches exist [18][16][12] for improving the blur kernel estimation step, but they assume only translational camera motion, which significantly restricts their utility. Several extensions consider rotations $r_x$,$r_y$ and $r_z$ to parametrize the 3D motion of the camera. Traditional deblurring approaches made use of parametric blur models to estimate kernels. An exhaustive survey of blind deblurring algorithms can be found in \cite{lai2016comparative}.

% Conventional hand-designed formation models would require explicit estimation of all of these independent variables from a single blurred image, which is an extremely ill-posed problem. As a result, applying such algorithms on general dynamic scenes yields images with unpleasant artifacts and incomplete deblurring. 
% However, these methods are developed based on a rather strong constraint that scene is planar and blur arises from camera motion alone, which disregards the commonly occurring blur in practical settings. 

% To address this, [][] proposed methods to estimate patch-level linear kernels for deblurring dynamic scenes. Another class of algorithms need segmentation methods to help the deblurring of dynamic scenes [9,10,23]. However, these methods heavily depend on an accurate segmentation. Importantly, all of the above methods are not end-to-end and their image estimation iterative algorithms are computation and time-intensive as highly non-convex optimization problems need to be solved.

Recent works based on deep convolutional neural networks (CNN) have studied the benefits of replacing the image formation model with a parametric model that can be trained to emulate the non-linear relationship between blurred-sharp image pairs. Such works~\cite{nah2017deep} directly regress to deblurred image intensities and overcome the limited representative capability of variational methods in describing dynamic scenes. These methods can handle combined effects of camera motion and dynamic object motion and achieve state-of-the-art results on single image deblurring task. They have reached a respectable reduction in model size, but still lack in accuracy and are not real-time.
% In this paper, we address this critical and general issue, and design a new network structure. 

Existing CNN-based methods have two major limitations:
a) Weights of the CNN are fixed and spatially invariant which may not be optimal for different pixels in a dynamically blurred scene (e.g., sky vs. moving car pixels). This issue is generally tackled by learning a highly non-linear mapping by stacking a large number of filters. But this drastically increases the computational cost and memory consumption. %Thus, it is difficult to use a CNN with a small model size to approximate the dynamic scene deblurring problem with the spatially variant property.
b) A geometrically uniform receptive field is sub-optimal for the task of deblurring. Large image regions tend to be used to increase the receptive field even though the blur is small. This inevitably leads to a network with a large number of layers and a high computation footprint which slows down the convergence of the network.

% Thus, there is a need to develop an effective network with low latency and a large receptive field to restore clear images from blurred dynamic scenes. 

Reaching a trade-off between the inference-speed, receptive field and the accuracy of a network is a non-trivial task (see Fig. 1). Our work focuses on the design of efficient and interpretable filtering modules that offer a better accuracy-speed trade-off as compared to simple cascade of convolutional layers. We investigate motion-dependent adaptability within a CNN to directly address the challenges in single image deblurring. Since motion blur is inherently directional and different for each image instance, a deblurring network can benefit from adapting to the blur present in each input test image. We deploy content-aware modules which adjust the filter to be applied and the receptive field at each pixel. Our analysis shows that the benefits of these dynamic modules for the deblurring task are two-fold:
i) Cascade of such layers provides a large and dynamically adaptive receptive field. Directional nature of blur requires a directional receptive field, which a normal CNN cannot achieve within a small number of layers.
ii) It efficiently enables spatially varying restoration, since changes in filters and features occur according to the blur in the local region. No previous work has investigated incorporating awareness of blur-variation within an end-to-end single image deblurring model.

%{\color{magenta} there is a disconnect between these paragraphs. Need to mention some disadvantage of the previous learning based methods or simply point out that we are superior.}
%We propose a network to approach the task of video deblurring  which can handle diverse types of motion blur.    

% We make this attempt by introducing blur-aware modulation of features and filters during the process of deblurring. Our experiments embrace this choice and show that such networks are well suited for generalized scene deblurring. --> use somewhere else
Following the state of the art in deblurring, we adopt a multi-patch hierarchical design to directly estimate the restored sharp image. Instead of cascading along the depth, we introduce content-aware feature and filter transformation capability through a global-local attentive module and residual attention across layers to improve performance. These modules learn to exploit the similarity in the motion between different pixels within an image and are also sensitive to position-specific local context. 

\begin{figure}[t]
\begin{center}
\includegraphics[width=\linewidth,height = 0.7\linewidth]{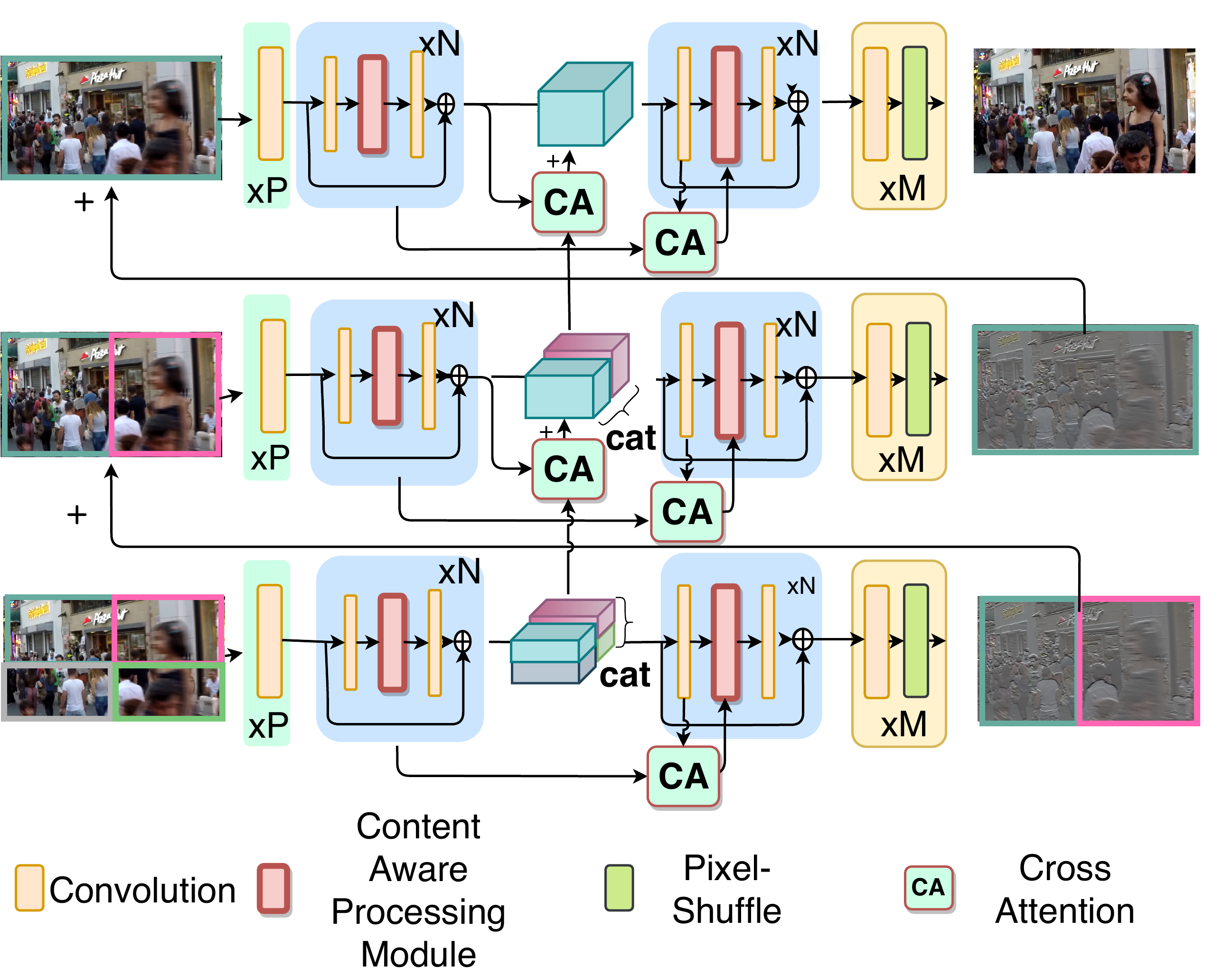}
\end{center}
   \caption{Overall architecture of our proposed network. CA block represents cross attention between different levels of encoder-decoder and different levels. All the resblock contains one content aware processing module. Symbol '+' denotes elementwise summation.}
\label{fig:mainarch}
\end{figure}

% It can generate local dynamic filters and global attention maps from the blurred image in a single shot in almost real-time.
% Our design originates from the intuition that motion blur is essentially an aggregation of various spatially varying transformations of the sharp regions, and hence a deblurring network can benefit from implicitly decoding the magnitude as well as the direction of motion accompanied by global context. To facilitate this, our proposed modules dynamically generate global attention maps and local filters. The feature transformations and filters estimated by the network are image dependent and hence can be visualized for different images. 
The efficiency of our architecture is demonstrated through a comprehensive evaluation on two benchmarks and comparisons with the state-of-the-art deblurring approaches. Our model achieves superior performance while being computationally more efficient. The major contributions of this work are:
\renewcommand{\labelenumi}{\Roman{enumi}}
\begin{itemize}

    \item We propose an efficient deblurring design built on new convolutional modules that learn the transformation of features using global attention and adaptive local filters. We show that these two branches complement each other and result in superior deblurring performance. Moreover, the efficient design of attention-module enables us to use it throughout the network without the need for explicit downsampling.
    
    \item We further demonstrate the efficacy of learning cross-attention between encode-decoder as well as different levels in our design.
    
    \item We provide extensive analysis and evaluations on dynamic scene deblurring benchmarks, demonstrating that our approach yields state-of-the-art results while being $3\times$ faster than the nearest competitor \cite{zhang2019deep}.
\end{itemize}

\section{Proposed Architecture}
To date, the driving force behind performance improvement in deblurring has been the use of a large number of layers and larger filters which assist in increasing the "static" receptive field and the generalization capability of a CNN. However, these techniques offer suboptimal design, since network performance does not always scale with network depth, as the effective receptive field of deep CNNs is much smaller than the theoretical value (investigated in \cite{luo2016understanding}).

We claim that a superior alternative is a dynamic framework wherein the filtering and the receptive field change across spatial locations and also across different input images. Our experiments show that this approach is a considerably better choice due to its task-specific efficacy and utility for computationally limited environments. It delivers consistent performance across diverse magnitudes of blur.

Although previous multi-scale and scale-recurrent methods have shown good performance in removing non-uniform blur, they suffer from expensive inference time and performance bottleneck while simply increasing model depth. Instead, inspired by \cite{zhang2019deep} , we adopt multi-patch hierarchical structure as our base-model, which compared to multi-scale approach has the added advantage of residual-like architecture that leads to efficient learning and faster processing speed.
%We build our multi-patch hierarchical architecture based upon the strong baseline of \cite{zhang2019deep} which is known to be superior to multi-scale designs  \cite{nah2017deep,tao2018scale} and achieves state-of-the-art performance. In contrast to multi-scale methods, this approach has the added advantage of residual-like architecture which leads to efficient learning and enables the use of smaller filters and hence faster processing.
The overall architecture of our proposed network is shown in Fig. \ref{fig:mainarch}. We divide the network into 3 levels instead of 4 as described in \cite{zhang2019deep}. We found that the relative performance gain due to the inclusion of level 4 is negligible compared to the increase in inference time and number of parameters. At the bottom level input sliced into 4 non-overlapping patches for processing, and as we gradually move towards higher levels, the number of patches decrease and lower level features are adaptively fused using attention module as shown in Fig. \ref{fig:mainarch}. The output of level 1 is the final deblurred image. Note that unlike \cite{zhang2019deep}, we also avoid cascading of our network along depth, as that adds severe computational burden. Instead, we advocate the use of content-aware processing modules which yield significant performance improvements over even the deepest stacked versions of original DMPHN \cite{zhang2019deep}. Major changes incorporated in our design are described next.

\par Each level of our network consists of an encoder and a decoder. Both the encoder and the decoder are made of standard convolutional layer and residual blocks where each of these residual blocks contains 1 convolution layer followed by a content-aware processing module and another convolutional layer. The content-aware processing module comprises two branches for global and local level feature processing which are dynamically fused at the end. The residual blocks of decoder and encoder are identical except for the use of cross attention in decoder. We have also designed cross-level attention for effective propagation of lower level features throughout the network. We begin with describing content-aware processing module, then proceed towards the detailed description of the two branches and finally how these branches are adaptively fused at the end.

\begin{figure*}[t]
\begin{center}
\includegraphics[width=0.8\linewidth,height=0.4\linewidth]{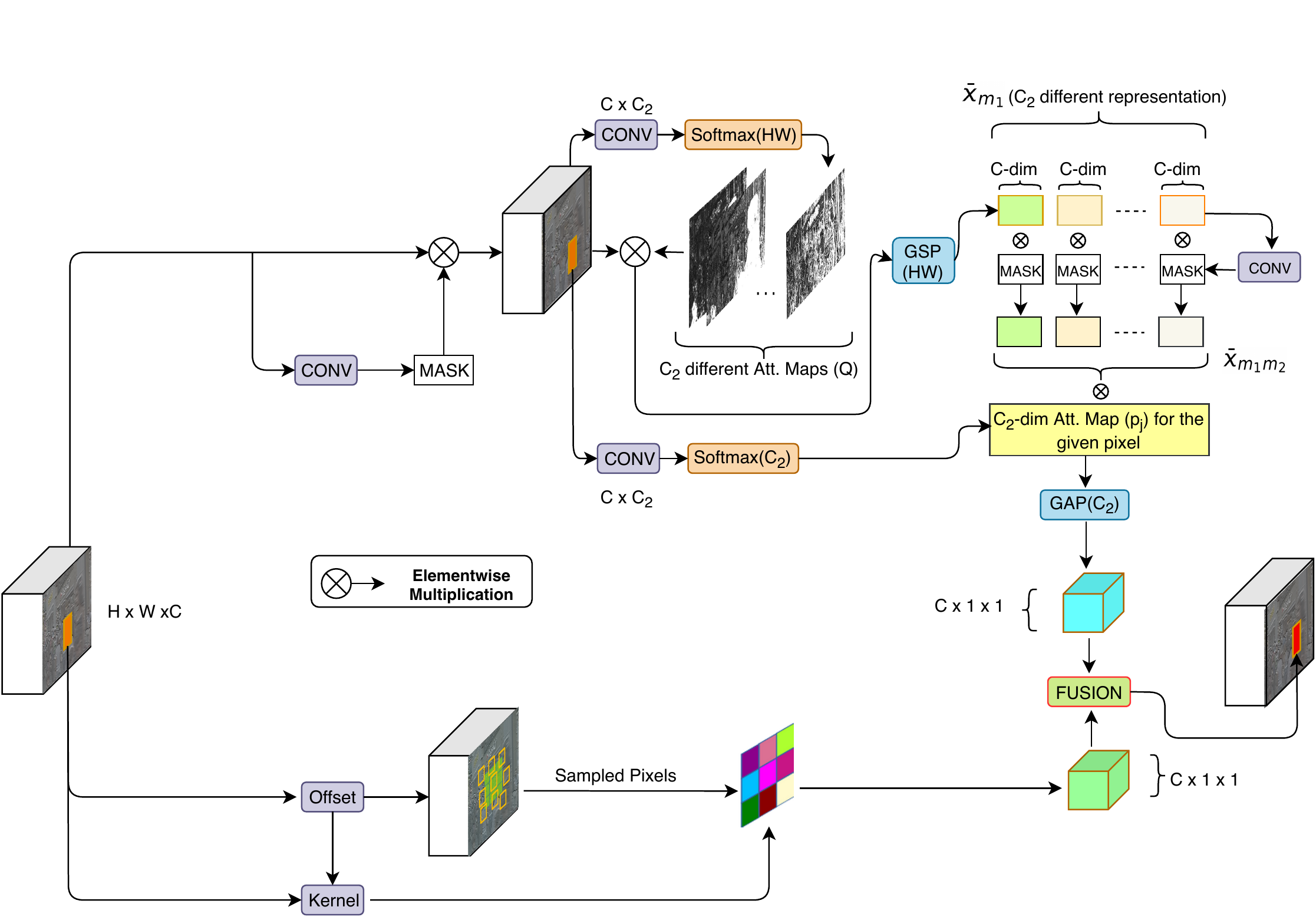}
\end{center}
   \caption{Illustration of our content-aware processing module. The upper and lower branch show self-attention (Sec. 3.1.1) and PDF module (Sec. 3.2). The fusion module is described in Eqs. 12 and 13.}
\label{fig:long}
\label{fig:onecol}
\end{figure*}

\section{Content-Aware Processing Module}
% Most of the existing CNN-based image restoration techniques focus mainly on designing a deeper or wider network to learn more discriminative high-level features, while rarely exploiting the inherent feature correlations in intermediate layers, thus hindering the representational ability of CNNs. Large receptive field is critical to the learning capacity of CNN. 
In contrast to high-level problems such as classification and detection \cite{wang2018non}, which can obtain large receptive field by successively down-sampling the feature map with pooling or strided convolution, restoration tasks like deblurring need finer pixel details that can not be achieved from highly downsampled features.
% On the other hand \cite{liu2018non} argued about the necessity of local neighbourhood for low level image restoration tasks and \cite{dai2019second} followed the same approach for image super-resolution. 
Most of the previous deblurring approaches uses standard convolutional layers for local filtering and stack those layers together to increase the receptive field. \cite{bello2019attention} uses self-attention and standard convolution on parallel branch and shows that best results are obtained when both features are combined together compared to using each feature separately. Inspired by this approach, we design a content-aware ``global-local'' processing module which depending on the input, deploys two parallel branches to fuse global and local features. The ``global'' branch is made of attention module. For decoder, this includes both self and cross-encoder-decoder attention whereas for encoder only self-attention is used. For local branch we design a pixel-dependent filtering module which determines the weight and the local neighbourhood to apply the filter adaptively. We describe in detail these two branches and their adaptive fusion strategy in the following sections.
\subsection{Attention}
Following the recent success of transformer architecture \cite{vaswani2017attention} in natural language processing domain, it has been introduced in image processing tasks as well \cite{parmar2018image,liu2018non}. The main building block of this architecture is self-attention which as the name suggests calculates the response at a position in a sequence by attending to all positions within the same sequence. Given an input tensor of shape $(C,H,W)$ it is flattened to a matrix $z \in \mathbb{R}^{HW \times C}$ and projected to $d_a$ and $d_c$ dimensional spaces using embedding matrices $W_a,W_b \in \mathbb{R}^{C \times d_a}$ and $W_c \in \mathbb{R}^{C \times d_c}$. Embedded matrices $A,B \in \mathbb{R}^{HW \times d_a}$ and $C \in \mathbb{R}^{HW \times d_c}$ are known as query, key and value, respectively. The output of the self-attention mechanism for a single head can be expressed as
\begin{equation}\label{eq:qkv}
    O = \text{softmax} \left(\frac{AB^T}{\sqrt{d_a}} \right)C
\end{equation}
The main drawback of this approach is very high memory requirement due to the matrix multiplication $AB^T$ which requires storing a high dimensional matrix of dimension $(HW,HW)$ for image domain. This requires a large downsampling operation before applying attention. \cite{parmar2018image} and \cite{ramachandran2019stand} use a local memory block instead of global all-to-all for making it practically usable. \cite{bello2019attention} uses attention only from the layer with the smallest spatial dimension until it hits memory constraints. Also, these works typically resort to smaller batch size and sometimes additionally downsampling the inputs to self-attention layers. Although self attention is implemented in recent video super-resolution work \cite{yi2019progressive}, to reduce memory requirement it resorts to pixel-shuffling. This process is sub-optimal for spatial attention as pixels are transferred to channel domain to reduce the size.
\par Different from others, we resort to an attention mechanism which is lightweight and fast. If we consider Eq. (\ref{eq:qkv}) without the softmax and scaling factor for simplicity, we first do a $(HW,d_a) \times (d_a,HW)$ matrix multiplication and then another $(HW,HW) \times (HW,d_c)$ matrix multiplication which is responsible for the high memory requirement and has a complexity of $\mathcal{O}(d_a(HW)^2)$. Instead, if we look into this equation differently and first compute $B^TC$ which is an $(d_a,HW) \times (HW,d_c)$ matrix multiplication followed by $A(B^TC)$ which is an $(HW,d_a) \times (d_a,d_c)$ matrix multiplication, this whole process becomes lightweight with a complexity of $\mathcal{O}(d_ad_cHW)$. We suitably introduce softmax operation at two places which makes this approach intuitively different from standard self-attention but still efficiently gathers global information for each pixel. Empirically we show that it performs better than standard self-attention as discussed in ablation studies. Also due to the light-weight nature, it not only enables us to use this in all the encoder and decoder blocks across levels for self-attention but also across different layers of encoder-decoder and levels for cross attention which results in a significant increase of accuracy. 
% The whole attention process can be broadly divided in 3 parts: a) Enhancing important features with motion information b) Aggregating global spatial feature in each pixel c) Aggregating global channel feature in each pixel. We describe each step in detail next.
\subsubsection{Self-Attention (SA)}
We start with generating a spatial attention mask $M_1$ describing which spatial features to emphasize or suppress for better motion understanding. Given the input feature map $x\in\mathbb{R}^{C \times H \times W}$ we generate $M_1$ as
\begin{equation}
    M_1 = f_{m_1}(x;\theta_{m_1})
\end{equation}
where $M_1 \in \mathbb{R}^{H \times W}$, $f_{m_1}(\cdot)$ is convolution followed by a sigmoid operation to generate a valid attention map. We generate the enhanced feature map $x_{m_1}$ by element-wise multiplication as
\begin{equation}\label{eq:sa1}
    x_{m_1} = x \odot M_1
\end{equation}
where $x_m \in \mathbb{R}^{C \times H \times W}$ and $M$ is broadcast along channel dimension accordingly. Next, we distribute these informative features to all the pixels adaptively which is similar to standard self-attention operation.

\par Given $x_m$, we generate three attention maps $P \in \mathbb{R}^{C_2 \times HW}$ , $Q \in \mathbb{R}^{C_2 \times HW}$ and $M_2 \in \mathbb{R}^{C}$ using convolutional operations $f_p(\cdot)$ ,$f_q(\cdot)$ and $f_{M_2}(\cdot)$ where global-average-pooling is used for the last case to get $C$ dimensional representation.  We take the first cluster of attention map $Q$ and split it into $C_2$ different maps $Q =\{q_1,q_2,...,q_{C_2}\}$, $q_i \in \mathbb{R}^{HW}$ and these represent $C_2$ different spatial attention-weights. A single attention reflects one aspect of the blurred image. However, there are multiple pertinent properties like edges,textures etc. in the image  that together helps removing the blur. Therefore, we deploy a cluster of attention maps to effectively gather $C_2$ different key features. Each attention map is element-wise multiplied with the input feature map $x_{m_1}$ to generate $C_2$ part feature maps as
\begin{equation}
    x^k_{m_1} = q_k \odot x_{m_1} ~ ~ ~ , \text{with}  \sum_{i=1}^{HW} q_{ki} = 1 ~ ~ ~ ~ ~ ~(k= 1,2,...,N)
\end{equation}
where $x^k_m \in \mathbb{R}^{C \times HW}$. We further extract descriptive global feature by global-sum-pooling (GSP) along $HW$ dimension to obtain $k^{th}$ feature representation as 
\begin{equation}
    \bar{x}^k_{m_1} = GSP_{HW}(x^k_{m_1}) ~ ~ ~ ~ ~ ~(k= 1,2,...,N)
\end{equation}
where $\bar{x}^k_m \in \mathbb{R}^C$. Now we have $\bar{x}_{m_1} = \{\bar{x}^1_{m_1},\bar{x}^2_{m_1},...,\bar{x}^{C_2}_{m_1}\}$ which are obtained from $C_2$ different attention-weighted average of the input $x_m$. Each of these $C_2$ representations is expressed by an $C$-dimensional vector which is a feature descriptor for the $C$ channels. Similar to the first step (Eq.(\ref{eq:sa1})), we further enhance these $C$ dimensional vectors by emphasizing the important feature-embeddings as
\begin{equation}\label{eq:sa2}
    \bar{x}^k_{{m_1}{m_2}} = M_2 \odot \bar{x}^k_{m_1}
\end{equation}
where $M_2$ can be expressed as
\begin{equation}
     M_2 = f_{m_2}(\bar{x}_{m_1};\theta_{m_2}) \in \mathbb{R}^C
\end{equation}
Eq.(\ref{eq:sa1}) and Eq.(\ref{eq:sa2}) can be intuitively compared to \cite{huang2019attention}, where similar gated-enhancement technique is used to refine the result by elementwise-multiplication with an attention mask that helps in propagating only the relevant information. Next we take the set of attention maps $P = \{p_1,p_2,...,p_{HW}\}$ where $p_i \in \mathbb{R}^{C_2}$ is represents attention map for $i^{th}$ pixel. Intuitively, $p_i$ shows the relative importance of $C_2$ different attention-weighted average ($\bar{x}_{{m_1}{m_2}}$) for the current pixel and it allows the pixel to adaptively select the weighted average of all the pixels. For each output pixel $j$, we element-wise multiply these $C_2$ feature representations $\bar{x}^k_{{m_1}{m_2}}$ with the corresponding attention map $p_j$, to get
\begin{equation}
y^j = p_j \odot \bar{x}_{{m_1}{m_2}} ~ ~ \text{with}  \sum_{i=1}^{C_2} p_{ji} = 1 ~ ~ , (j= 1,2,...,HW)
\end{equation}
where ${y}^j \in \mathbb{R}^{C \times C_2}$. We again apply global-average-pooling on ${y}^j$ along $C_2$ to get $C$ dimensional feature representation for each pixel as
\begin{equation}
    \bar{y}^{j} = GAP_{C_2}({y}^j)
\end{equation}
where $\bar{y}^{j} \in \mathbb{R}^C$ represent the accumulated global feature for the $j^{th}$ pixel. Thus, each pixel flexibly selects features that are complementary to the current one and accumulates a global information.
This whole sequence of operations can be expressed by efficient matrix-operations as
% \begin{equation}
%     % x^{att} = \left[f_a(x_m)softmax(f_b(x_m))^T\right]softmax(f_c(x_m))
%     y^{att} = 
% \end{equation}

\begin{equation}
    y^{att} =  C\odot \left[(A)\text{softmax}(B)^T\right]\text{softmax}(D) \\ 
\end{equation} 
where $A$, $B$, $C$, $D$ are given by
\begin{gather*}
    C = \sigma (f_{M_2}(x_{m_1})) \in  \mathbb{R}^{C}  , A = \sigma(f_{M_1}(x)) \in  \mathbb{R}^{C \times HW} \\ 
    B = f_Q(x_{m_1}) \in  \mathbb{R}^{HW \times C_2}, D = f_P(x_{m_1}) \in  \mathbb{R}^{C_2 \times HW}
\end{gather*}
This efficient and simple matrix multiplication makes this attention module very fast whereas the order of operation (first computing $[(A)\text{softmax}(B)^T]$) results in low memory footprint. Note that, $C$ is broadcast along $HW$ dimension appropriately. We utilize this attention block in both encoder and decoder at each level for self-attention.
\subsubsection{Cross-Attention (CA)}
Inspired from the use of cross-attention in \cite{vaswani2017attention}, we implement cross encoder-decoder and cross level attention in our model. For cross encoder-decoder attention, we deploy similar attention module where the information to be attended is from different encoder layers and all the attention maps are generated by the decoder. Similarly for cross-level, the attended feature is from a lower level and the attention decisions are made by features from a higher level. We have observed that this helps in the propagation of information across layers and levels compared to simply passing the whole input or doing elementwise sum as done in \cite{zhang2019deep}. 
%cross-module attention is between encoder and decoder at each level and cross-level attention is between encoded features of different levels which helps in the propagation of information across modules and levels.

\begin{figure*}[htb] \label{fig:visual_deblur}
	\scriptsize
	\centering
		%\tiny
			\begin{tabular}{cccccccc}
\includegraphics[width=0.10\textwidth]{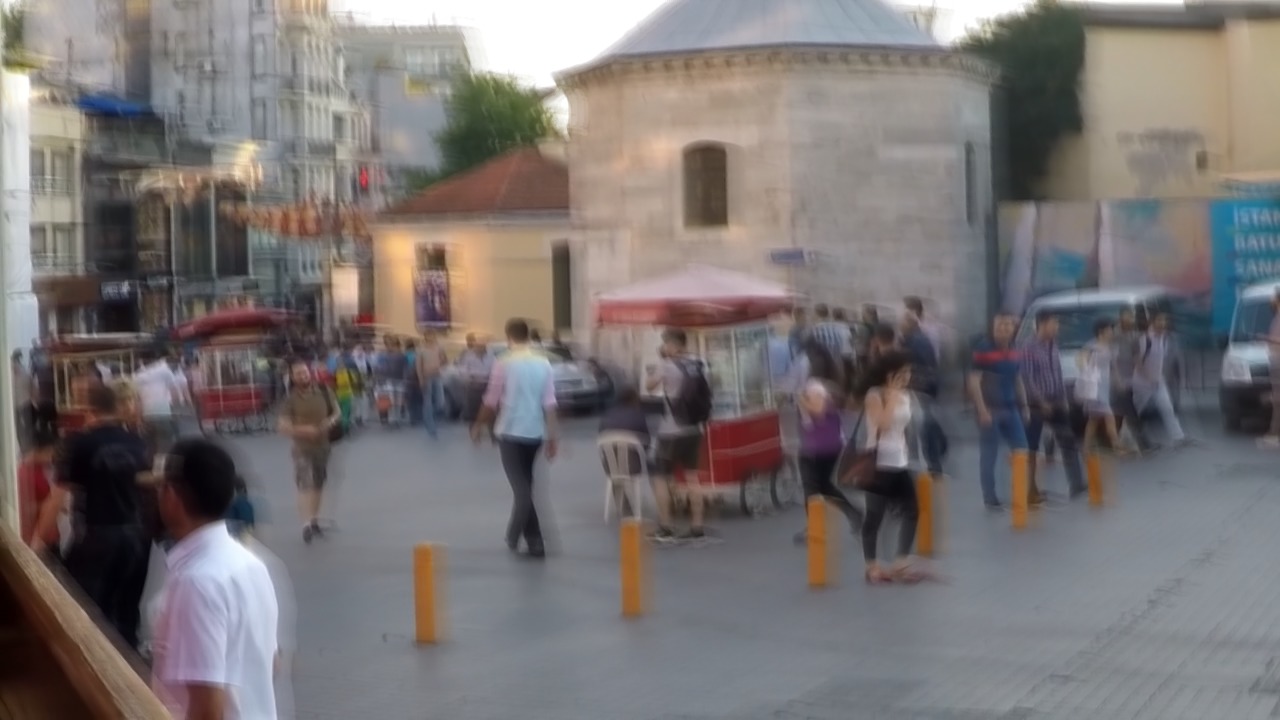} &
		        \includegraphics[width=\widthscalefive \textwidth]{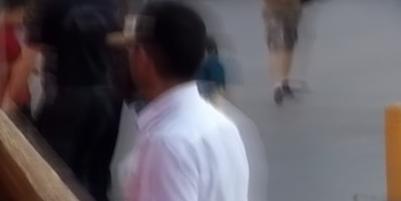} & 
				\includegraphics[width=\widthscalefive \textwidth]{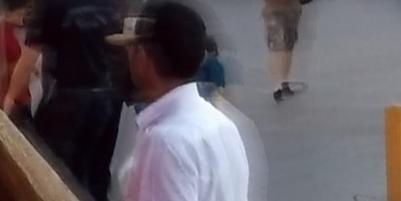} & %\hspace{\fsdttwofig} &
				\includegraphics[width=\widthscalefive \textwidth]{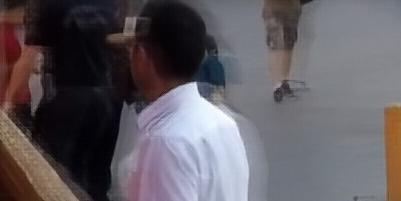} & %\hspace{\fsdttwofig} &
				\includegraphics[width=\widthscalefive \textwidth]{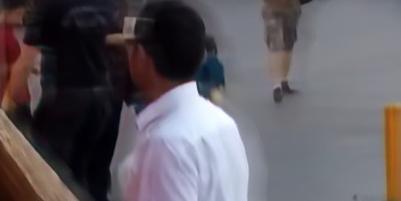} &
				\includegraphics[bb=30 70 440 275,clip=True,width=\widthscalefive \textwidth]{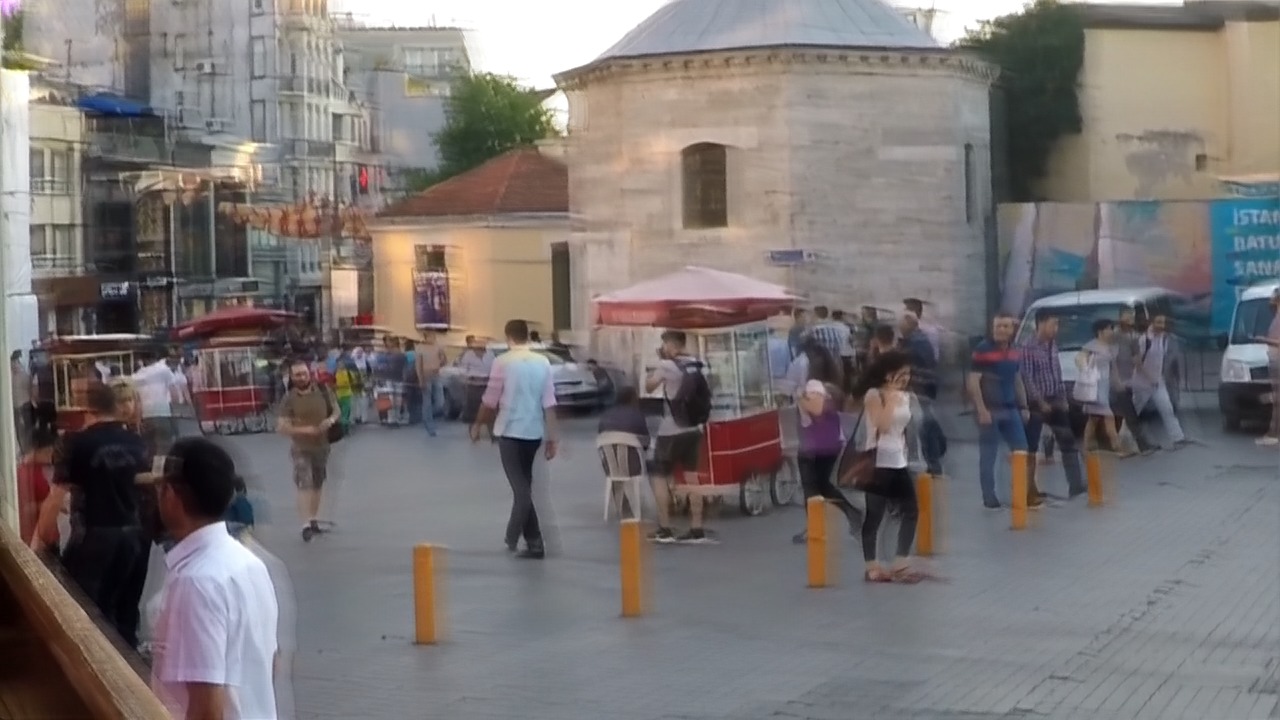} &
				 \includegraphics[width=\widthscalefive \textwidth]{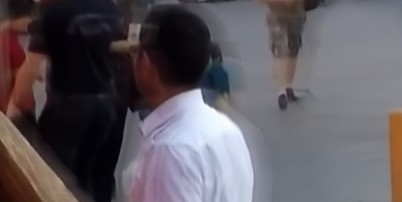} & %\hspace{\fsdttwofig} &
				\includegraphics[bb=30 70 440 275,clip=True,width=\widthscalefive \textwidth]{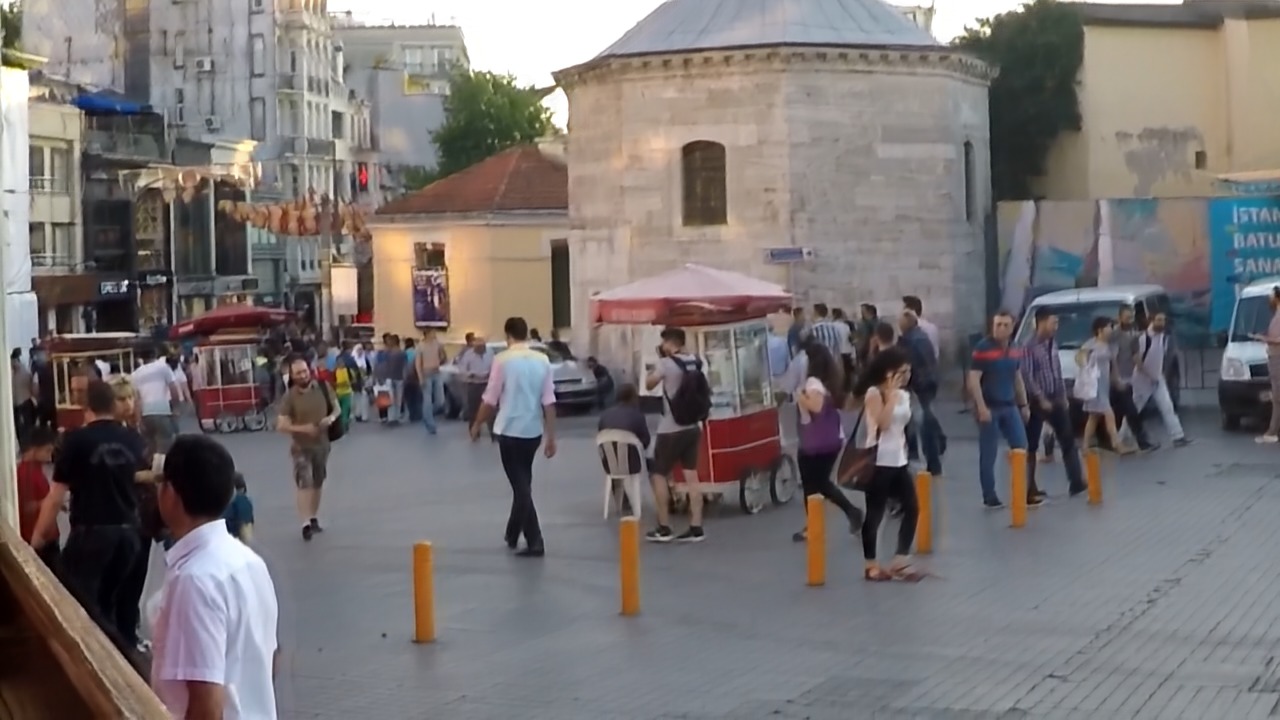}
				\\
				\includegraphics[width=0.10\textwidth]{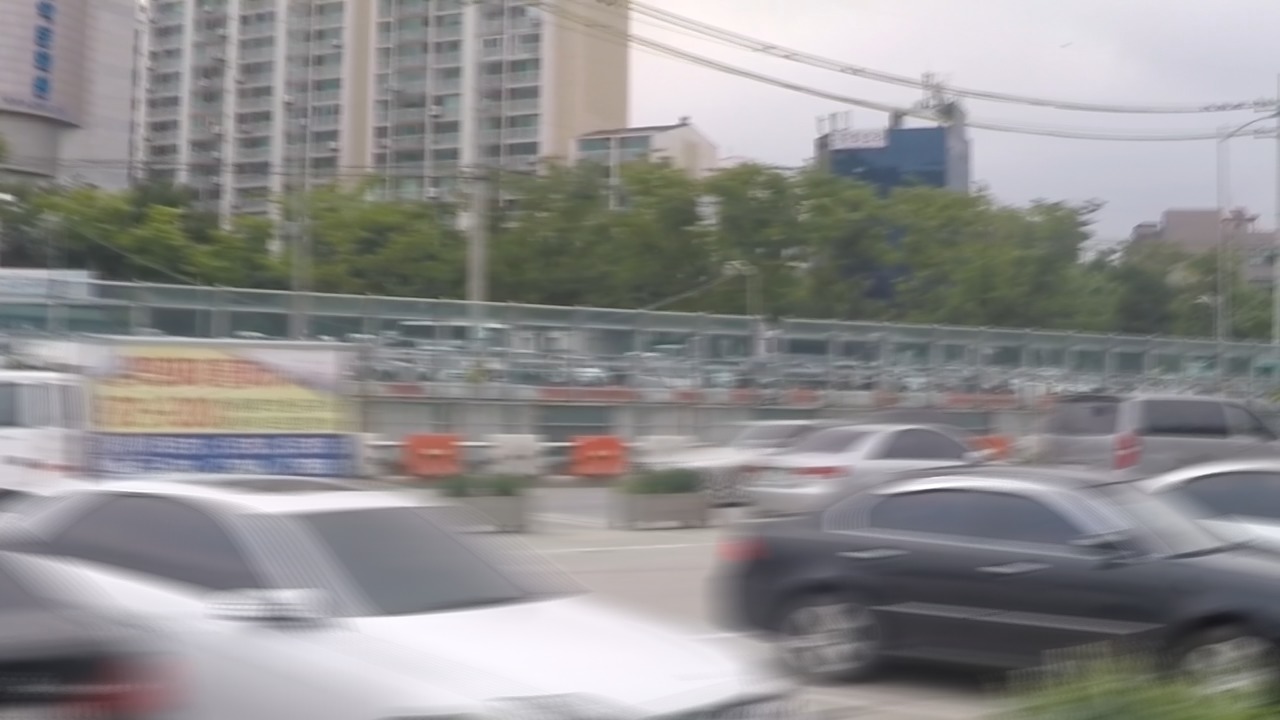}
				&
				\includegraphics[width=\widthscalefive \textwidth]{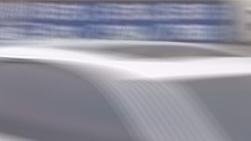} & %\hspace{\fsdttwofig} &

				\includegraphics[width=\widthscalefive \textwidth]{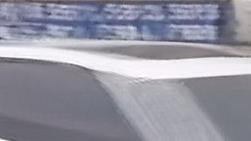} & %\hspace{\fsdttwofig} &
				\includegraphics[width=\widthscalefive \textwidth]{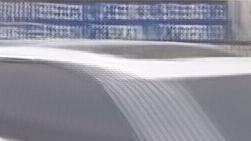} & %\hspace{\fsdttwofig} &
				\includegraphics[width=\widthscalefive \textwidth]{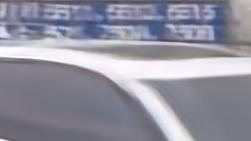} &
				\includegraphics[bb=80 130 390 300,clip=True,width=\widthscalefive \textwidth]{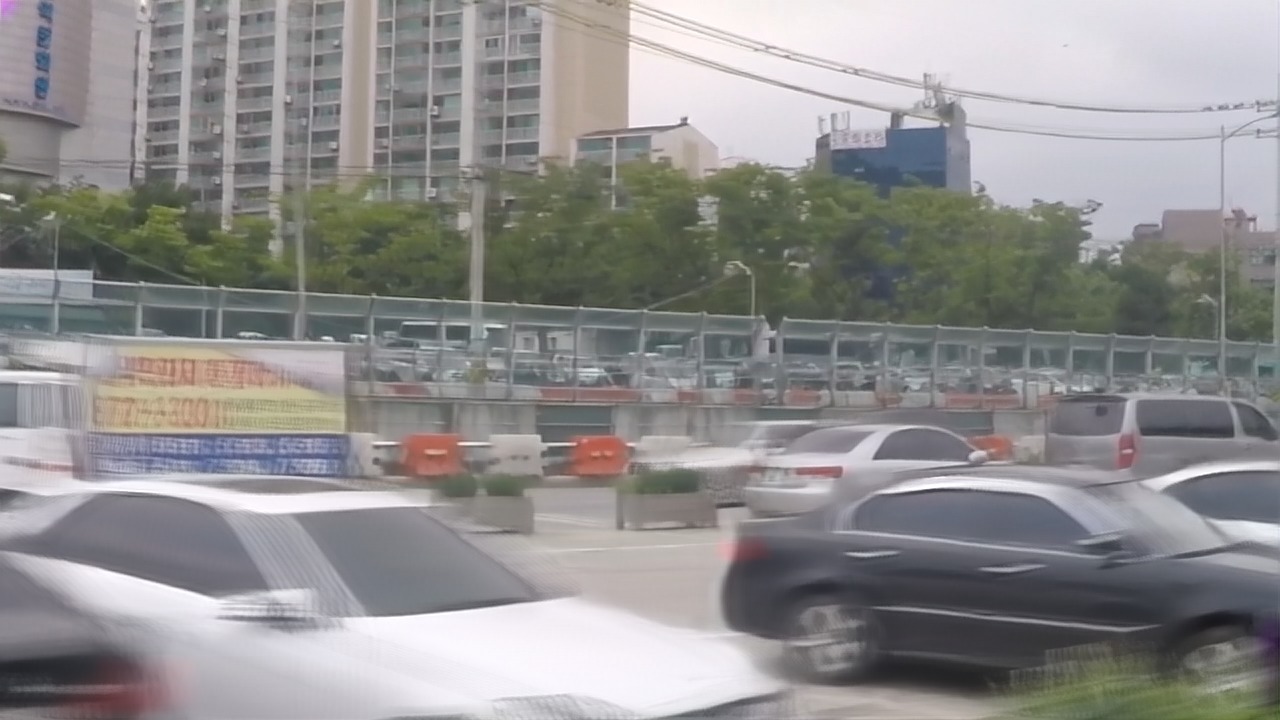} &
				\includegraphics[width=\widthscalefive \textwidth]{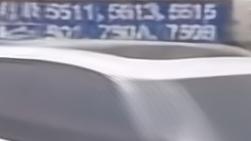} &				 %\hspace{\fsdttwofig} &3
				\includegraphics[bb=80 130 390 300,clip=True,width=\widthscalefive \textwidth]{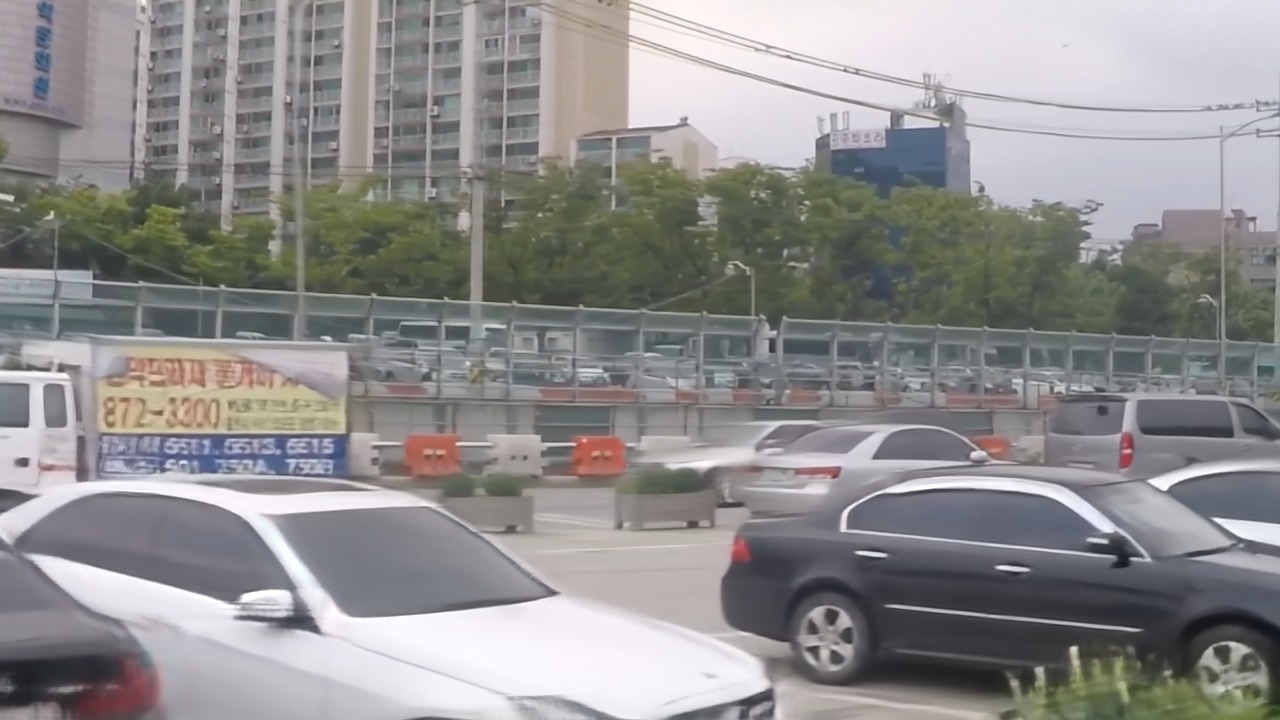}				%\vspace{-3mm}
\\
\\
				\includegraphics[width=0.10\textwidth]{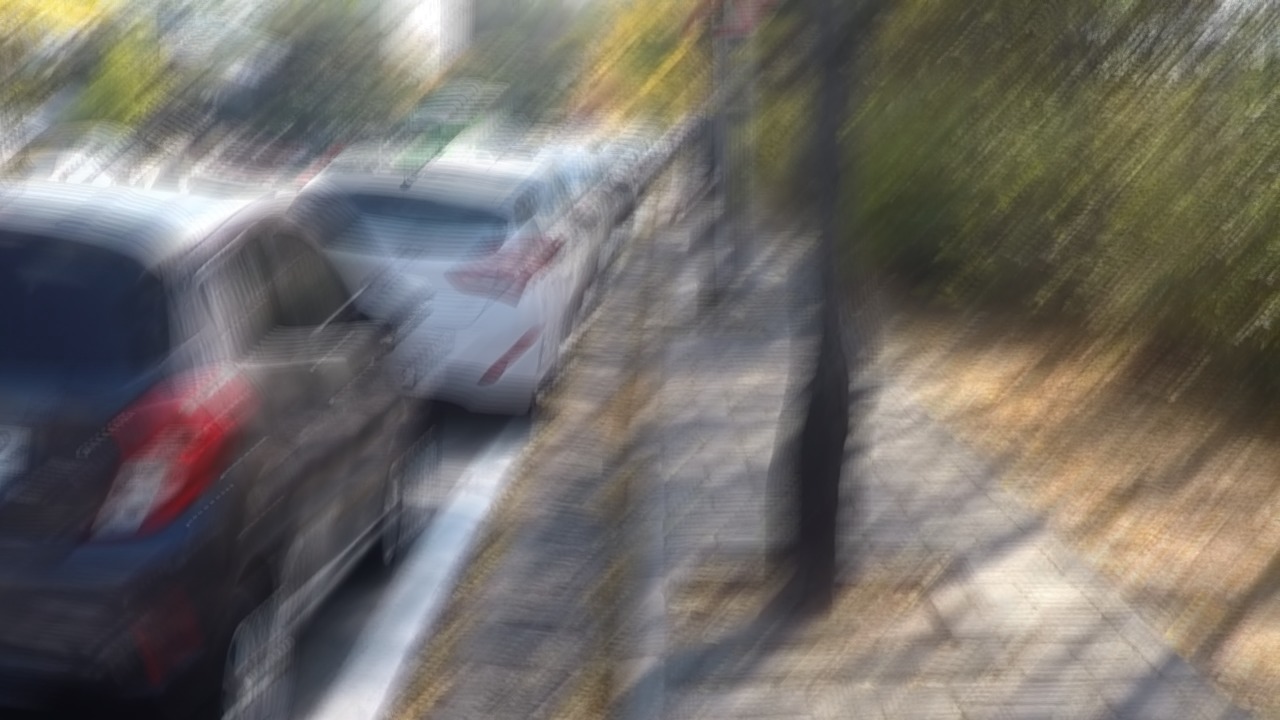}
				&
				\includegraphics[width=\widthscalefive \textwidth]{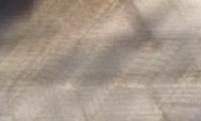} & %\hspace{\fsdttwofig} &

				\includegraphics[width=\widthscalefive \textwidth]{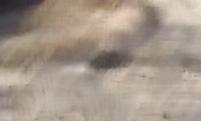} & %\hspace{\fsdttwofig} &
				\includegraphics[width=\widthscalefive \textwidth]{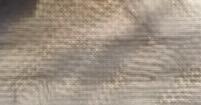} & %\hspace{\fsdttwofig} &
				\includegraphics[width=\widthscalefive \textwidth]{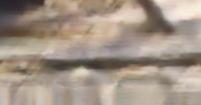} &
				\includegraphics[bb=820 20 1000 125,clip=True,width=\widthscalefive \textwidth]{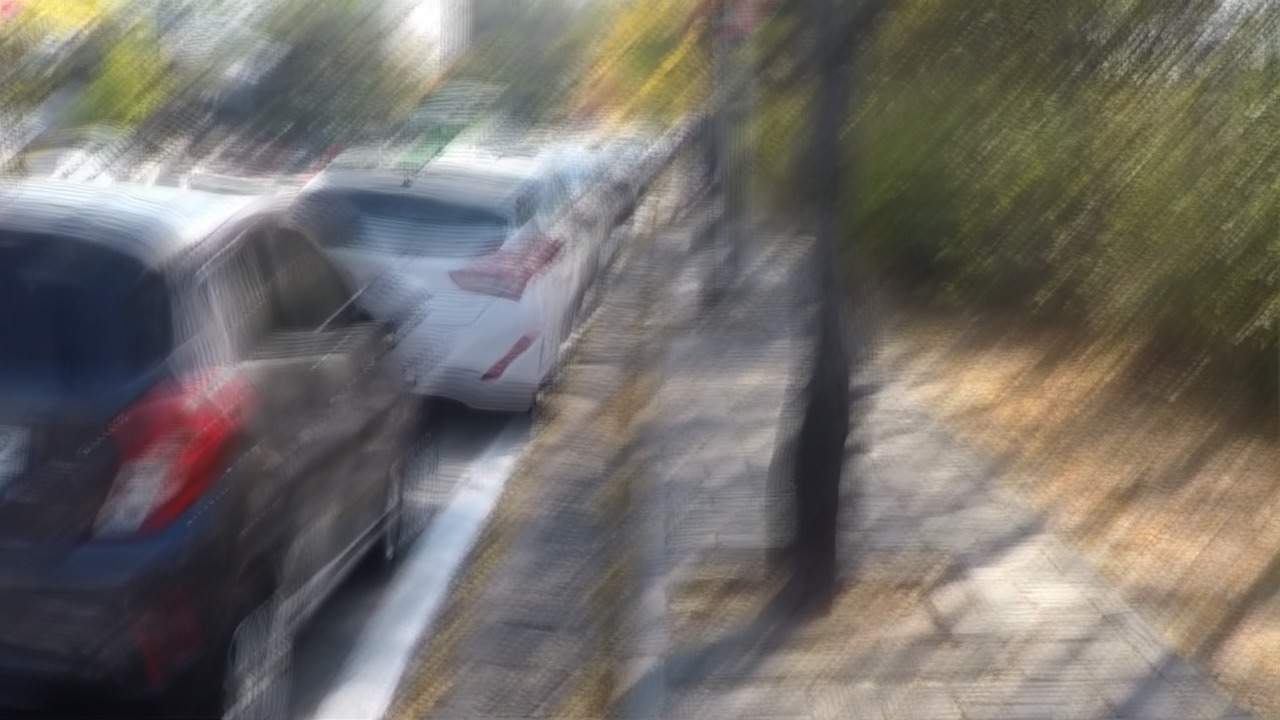} &
				\includegraphics[width=\widthscalefive \textwidth]{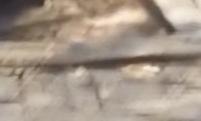} & 				%\hspace{\fsdttwofig} &3
				\includegraphics[bb=820 20 1000 125,clip=True,width=\widthscalefive \textwidth]{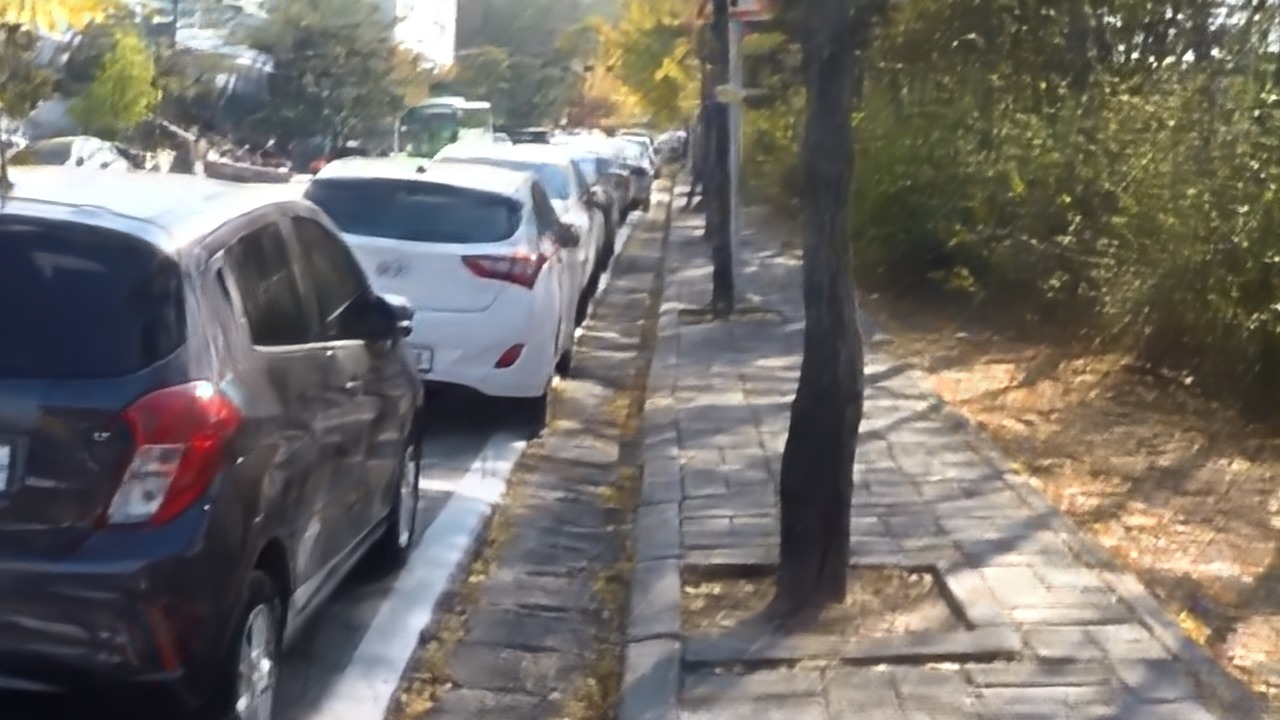}				%\vspace{-3mm}				
				 
				\\ 
				\includegraphics[width=0.10\textwidth]{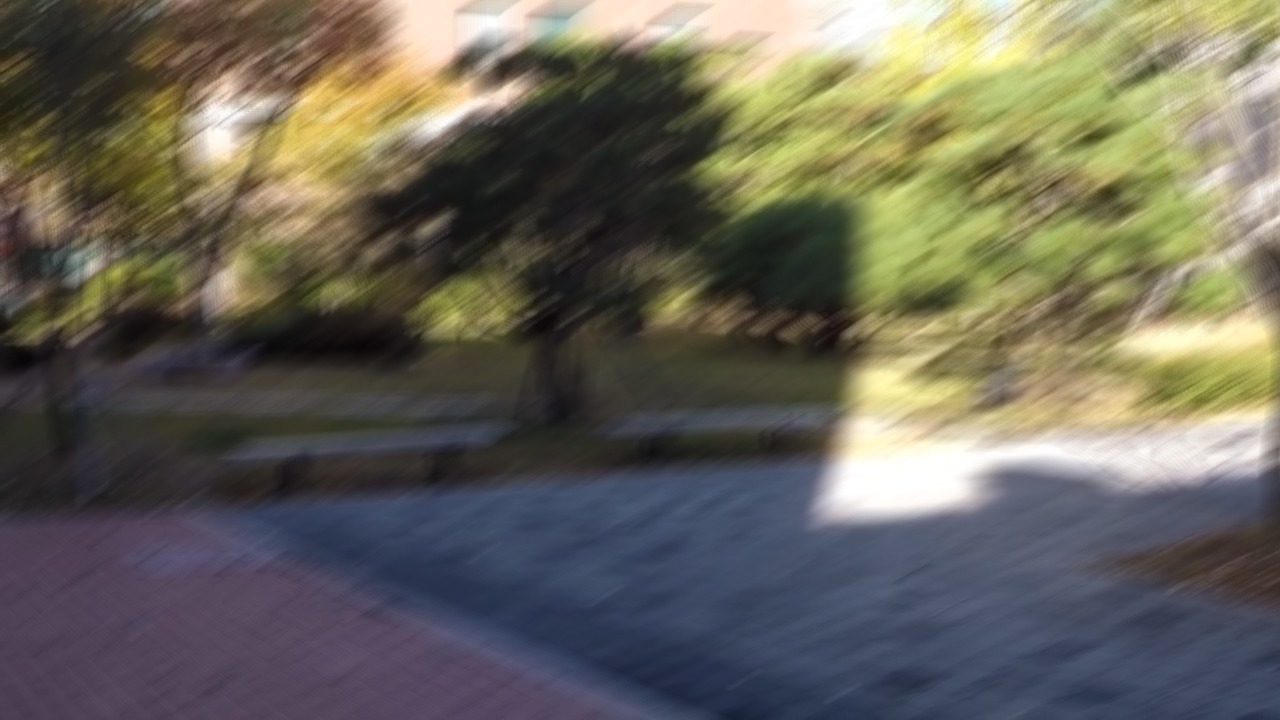}
				&
				\includegraphics[width=\widthscalefive \textwidth]{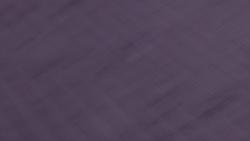} & %\hspace{\fsdttwofig} &

				\includegraphics[width=\widthscalefive \textwidth]{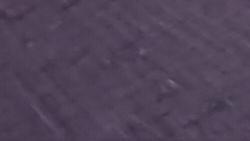} & %\hspace{\fsdttwofig} &
				\includegraphics[width=\widthscalefive \textwidth]{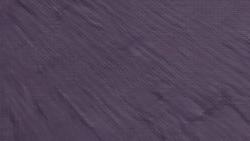} & %\hspace{\fsdttwofig} &
				\includegraphics[width=\widthscalefive \textwidth]{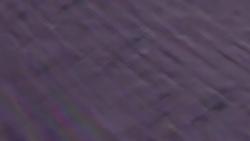} &
				\includegraphics[bb=10 0 300 165,clip=True,width=\widthscalefive \textwidth]{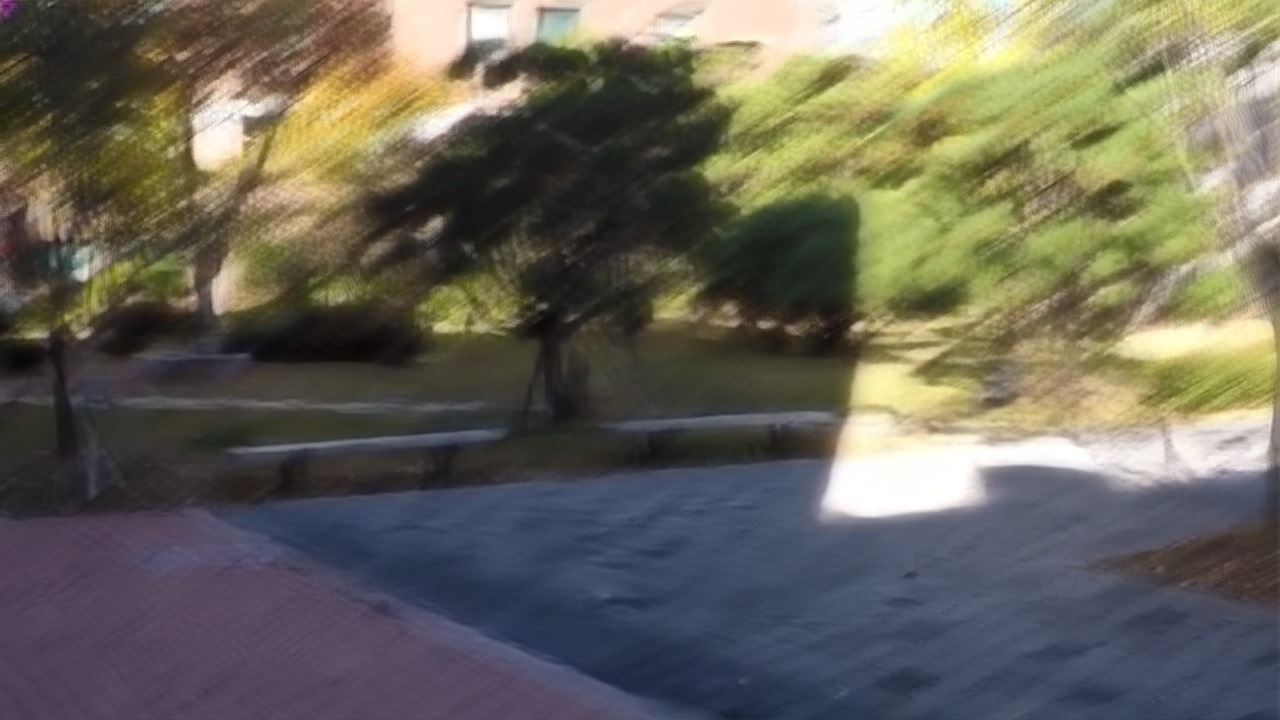} &
				\includegraphics[width=\widthscalefive \textwidth]{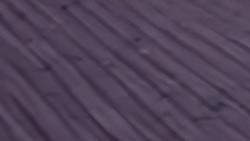} & 				%\hspace{\fsdttwofig} &3
				\includegraphics[bb=10 0 300 165,clip=True,width=\widthscalefive \textwidth]{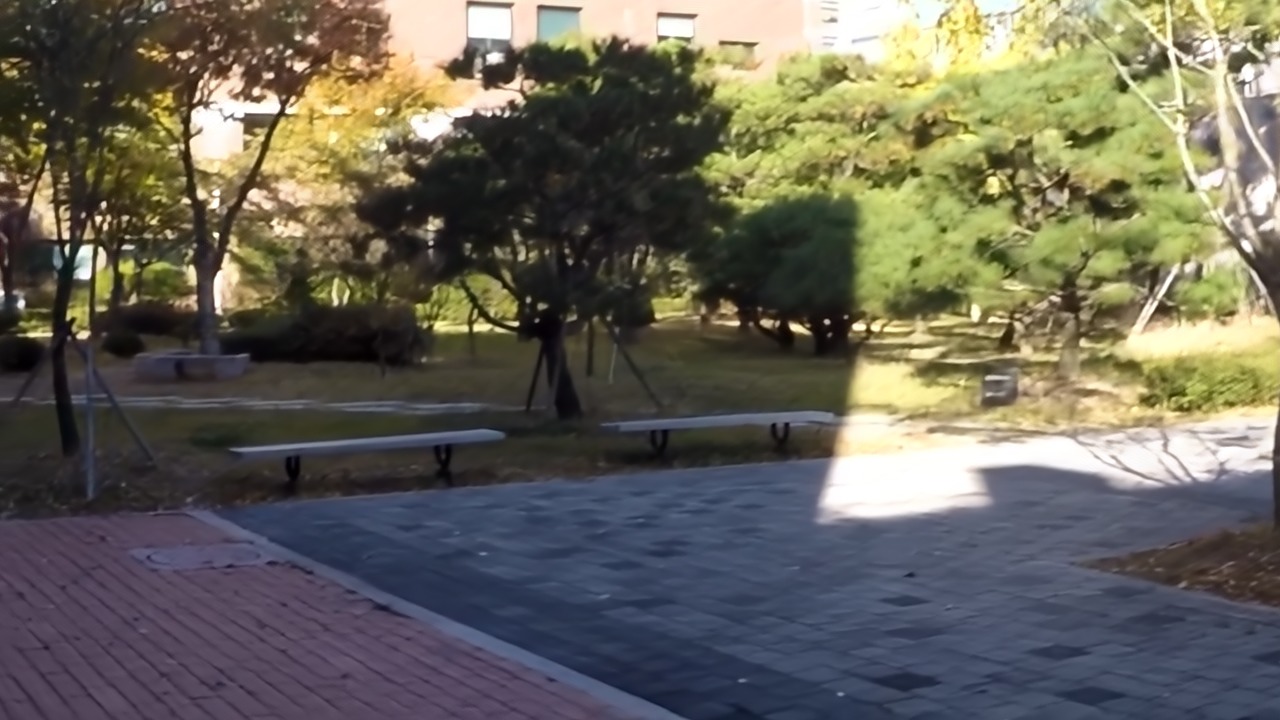}				%\vspace{-3mm}				
				 
				\\ 				
								(a) Blurred Image&

				(b) Blurred patch& %\hspace{\fsdttwofig} &
				(c) MS-CNN & %\hspace{\fsdttwofig} &
				(d) DelurGAN& % \hspace{\fsdttwofig} &
				(e) SRN & %\hspace{\fsdttwofig} &
				(f) DelurGAN-V2 & % \hspace{\fsdttwofig} &
				(g) \tiny{Stack(4)-DMPHN} & %\hspace{\fsdttwofig} &
				(h) Ours (a)%\hspace{\fsdttwofig} &
				\\
	\end{tabular}
	%\vspace{-0em}
	\caption{Visual comparisons of deblurring results on images from the GoPro test set~\cite{nah2017deep}. Key blurred patches are shown in (b), while zoomed-in patches from the deblurred results are shown in (c)-(h).}% (best viewed in high resolution).}
\label{fig:dynamic}
	%\vspace{-0em}
\end{figure*} 

\subsection{Pixel-Dependent Filtering Module (PDF)}
In contrast to \cite{bello2019attention}, for the local branch, we use Pixel-Dependent Filtering Module to handle spatially-varying dynamic motion blur effectively. Previous works like \cite{jia2016dynamic} generate sample-specific parameters on-the-fly using a filter generation network for image classification. \cite{li2018video} uses input text to construct the motion-generating filter weights for video generation task. \cite{zhang2018crowd} uses an adaptive convolutional layer where the convolution filter weights are the outputs of a separate “filter-manifold network” for crowd counting task. Our work is based on \cite{su2019pixel} as we use a ‘meta-layer’ to generate pixel dependent spatially varying kernel to implement spatially variant convolution operation. Along with that, the local pixels where the filter is to be applied, are also determined at runtime as we adjust the offsets of these filters adaptively. Given the input feature map $x \in  \mathbb{R}^{C \times H \times W}$, we apply a kernel generation function to generate a spatially varying kernel $V$ and do the convolution operation for pixel $j$ as
\begin{equation}
    y^{dyn}_{j,c}  = \sum_{k = 1}^{K} V_{j,j_k} W_c[j_k] x[j+j_k + \Delta j_k]
\end{equation}
where $y^{dyn}_j \in \mathbb{R}^C$, $K$ is the kernel size, $j_k \in \{(-(K-1)/2,-(K-1)/2),...,((K-1)/2,(K-1)/2)\}$ defines position of the convolutional kernel of dilation 1, $V_{j,j_k} \in \mathbb{R}^{{K^2} \times H \times W}$ is the pixel dependent kernel generated,$W_c \in \mathbb{R}^{C\times C\times K \times K}$ is the fixed weight and $\Delta j_k$ are the learnable offsets. We set a maximum threshold $\Delta_{\text{max}}$ for the offsets to enforce efficient local processing which is important for low level tasks like deblurring.  Note that the kernels ($V$) and offsets vary from one pixel to another, but are constant for all the channels, promoting efficiency. %This design choice makes our approach more efficient compared to \cite{zhou2019spatio}.
Standard spatial convolution can be seen as a special case of the above with adapting kernel being constant $V_{j,j_k} = 1$ and $\Delta j_k = 0$.
% Finally we sum-fuse the output of these two branches as
In contrast to \cite{bello2019attention}, which simply concatenates the output of these two branches, we design attentive fusion between these two branches so that the network can adaptively adjust the importance of each branch for each pixel at runtime. Empirically we observed that it performs better than simple addition or concatenation. Also, as discussed in visualization section, it gives an insight into the specific requirement for different levels of blur. Given the original input $x$ to this content-aware module, we generate a fusion mask as
\begin{equation}
    M_{fus} = sigmoid(f_{fus}(x))
\end{equation}
where $M_{fus} \in \mathbb{R}^{H \times W}$, $f_{fus}$ is a single convolution layer generating single channel output. Then we fuse the two branches as
\begin{equation}
    y^{GL} = M_{fus} \odot y^{att} + (1 - M_{fus}) \odot y^{dyn}
\end{equation}
The fused output $y^{GL}$ contains global as well as local information distributed adaptively along pixels which helps in handling spatially-varying motion blur effectively.

\begin{figure*}[htb] 
	\scriptsize
	\centering
		%\tiny
			\begin{tabular}{ccccccc}
\includegraphics[width=\widthscalesix\textwidth]{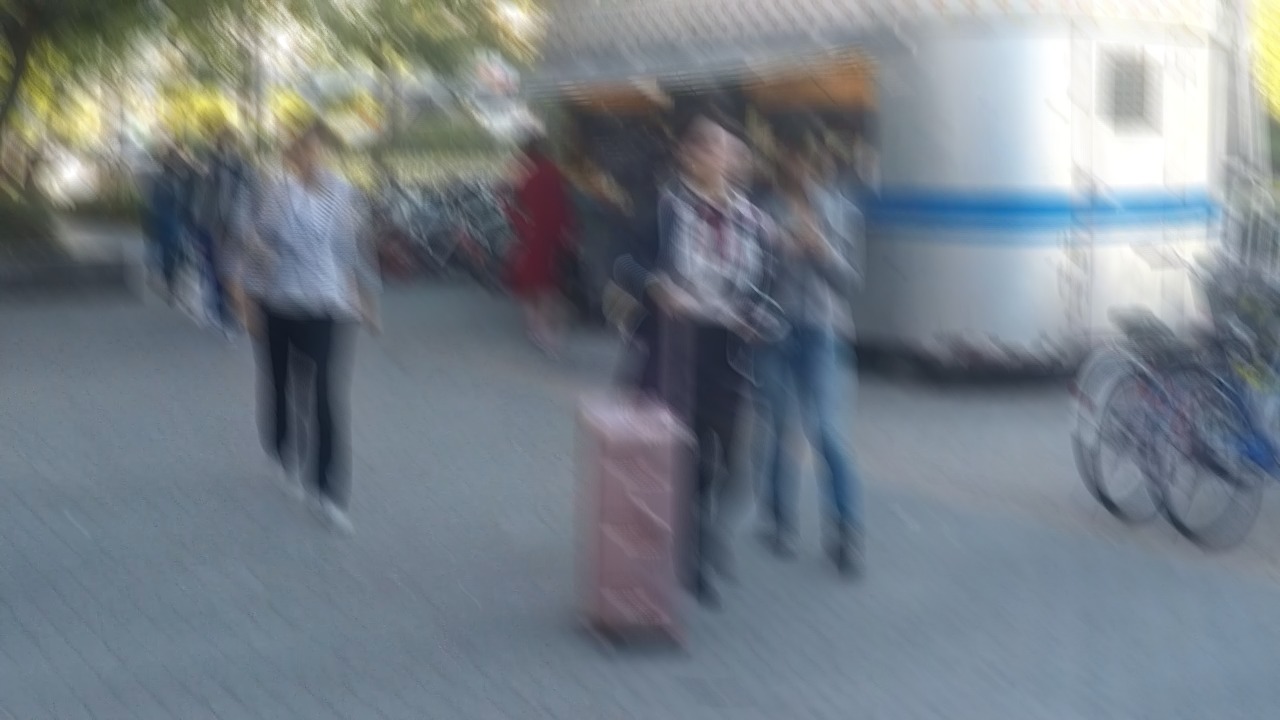} &
		        \includegraphics[width=\widthscalesix \textwidth]{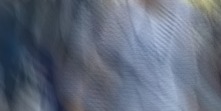} & 
				\includegraphics[width=\widthscalesix \textwidth]{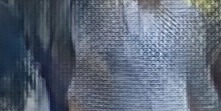} & %\hspace{\fsdttwofig} &
				\includegraphics[width=\widthscalesix \textwidth]{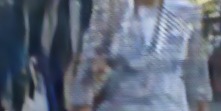} & %\hspace{\fsdttwofig} &
				 \includegraphics[width=\widthscalesix \textwidth]{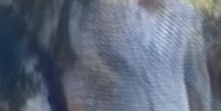} &
				 \includegraphics[width=\widthscalesix \textwidth]{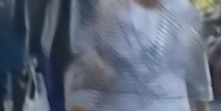} &				 %\hspace{\fsdttwofig} &
				\includegraphics[width=\widthscalesix \textwidth]{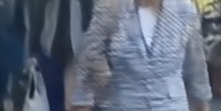}
				\\
\includegraphics[width=\widthscalesix\textwidth]{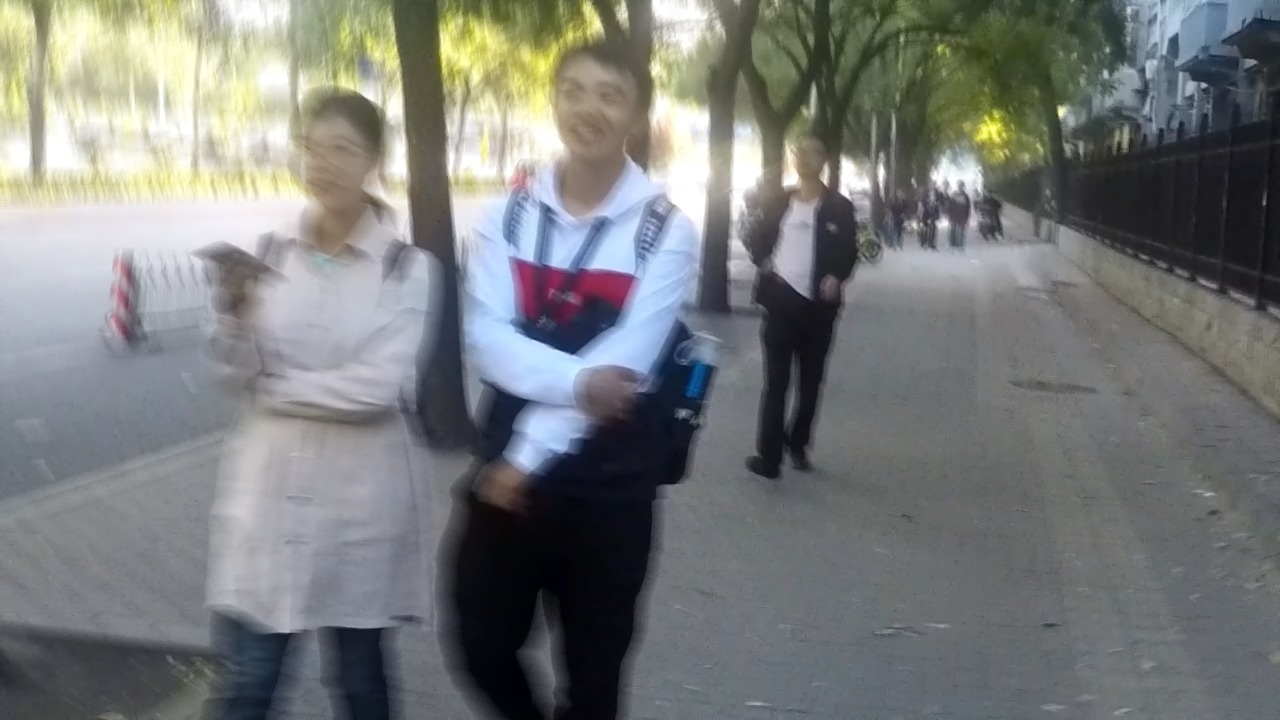} &
		        \includegraphics[width=\widthscalesix \textwidth]{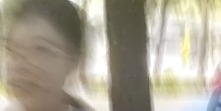} & 
				\includegraphics[width=\widthscalesix \textwidth]{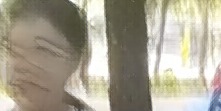} & %\hspace{\fsdttwofig} &
				\includegraphics[width=\widthscalesix \textwidth]{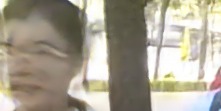} & %\hspace{\fsdttwofig} &
				 \includegraphics[width=\widthscalesix \textwidth]{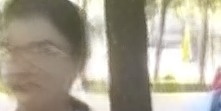} &
				 \includegraphics[width=\widthscalesix \textwidth]{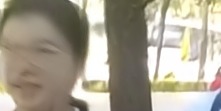} &				 %\hspace{\fsdttwofig} &
				\includegraphics[width=\widthscalesix \textwidth]{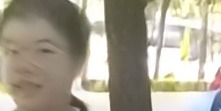}
\\
\includegraphics[width=\widthscalesix\textwidth]{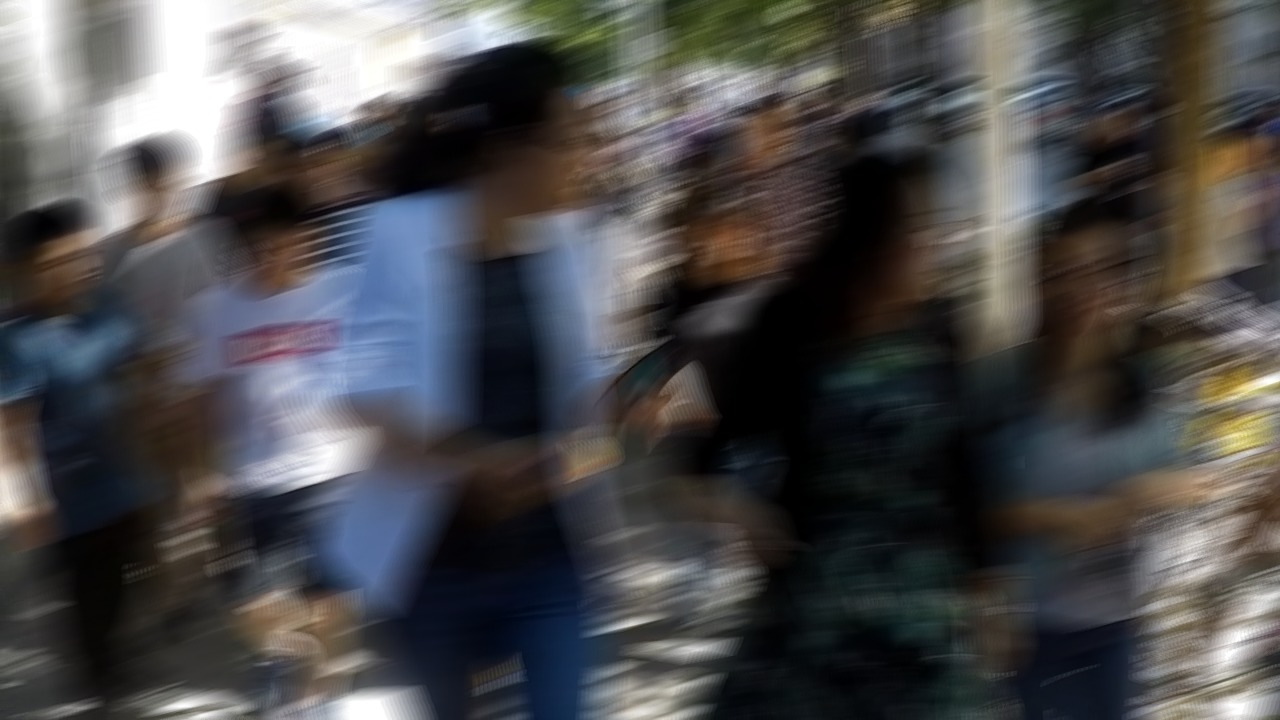} &
		        \includegraphics[width=\widthscalesix \textwidth]{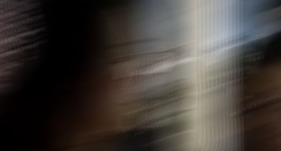} & 
				\includegraphics[width=\widthscalesix \textwidth]{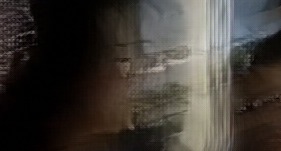} & %\hspace{\fsdttwofig} &
				\includegraphics[width=\widthscalesix \textwidth]{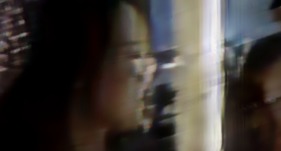} & %\hspace{\fsdttwofig} &
				 \includegraphics[width=\widthscalesix \textwidth]{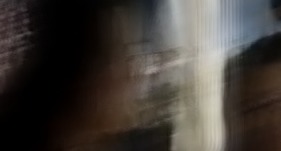} &
				 \includegraphics[width=\widthscalesix \textwidth]{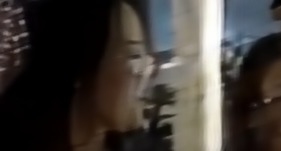} &				 %\hspace{\fsdttwofig} &
				\includegraphics[width=\widthscalesix \textwidth]{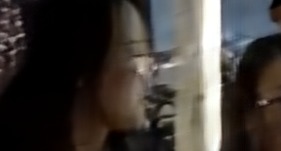}			
				 				\\ 
			
								(a) Blurred Image&

				(b) Blurred patch& %\hspace{\fsdttwofig} &
				(c) DelurGAN & %\hspace{\fsdttwofig} &
				(d) SRN & %\hspace{\fsdttwofig} &
				(e) DelurGANv2 & % \hspace{\fsdttwofig} &
				(f) \tiny{Stack(4)-DMPHN} & %\hspace{\fsdttwofig} &
				(g) Ours  %\hspace{\fsdttwofig} &
				\\
	\end{tabular}
	%\vspace{-0em}
	\caption{Visual comparisons of deblurring results on images from the HIDE test set~\cite{shen2019human}. Key blurred patches are shown in (b), while zoomed-in patches from the deblurred results are shown in (c)-(g).}% (best viewed in high resolution).}
\label{fig:visual_deblur_HIDE}
	%\vspace{-0em}
\end{figure*}

\begin{figure}[htb] \label{fusion}
	\centering
		%\tiny
			\begin{tabular}{cccc}
		        \includegraphics[width=0.3\linewidth]{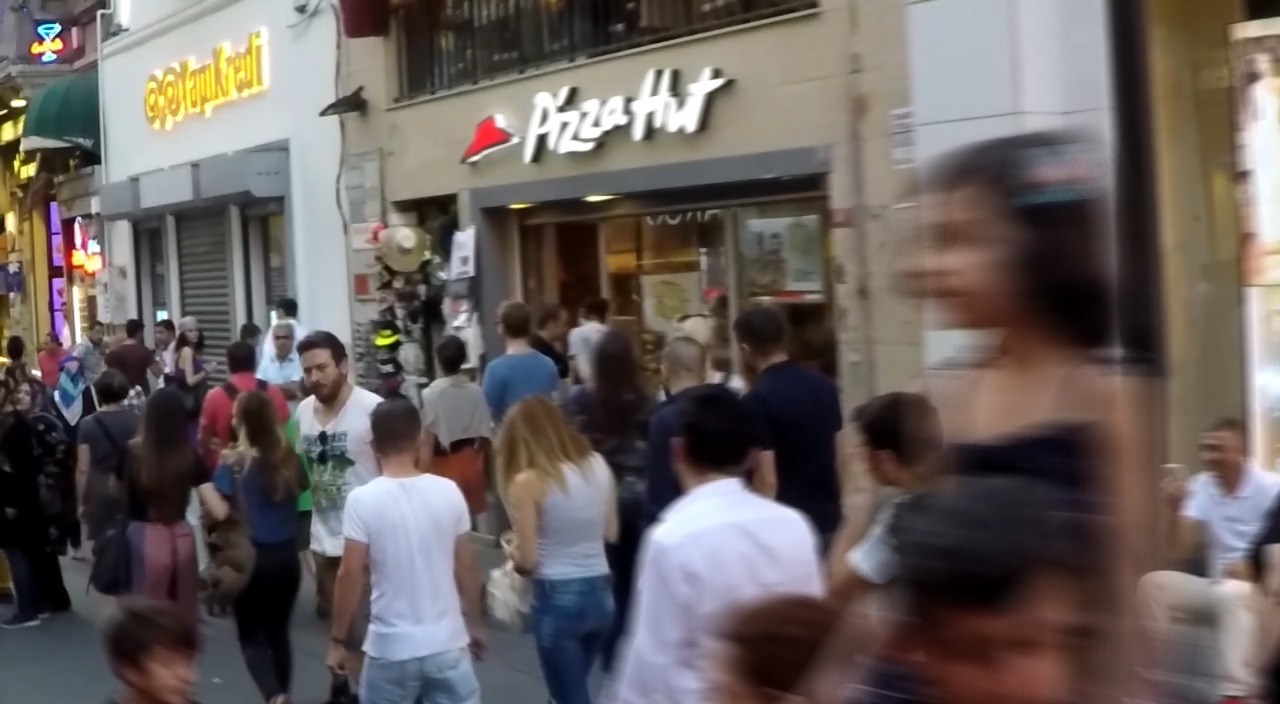} & 
				\includegraphics[bb=60 80 400 280,clip=True,width=0.3\linewidth]{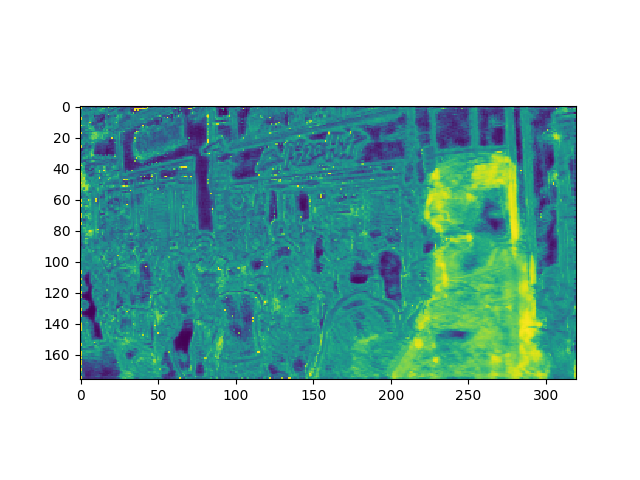} & %\hspace{\fsdttwofig} &
				\includegraphics[bb=60 80 400 280,clip=True,width=0.3\linewidth]{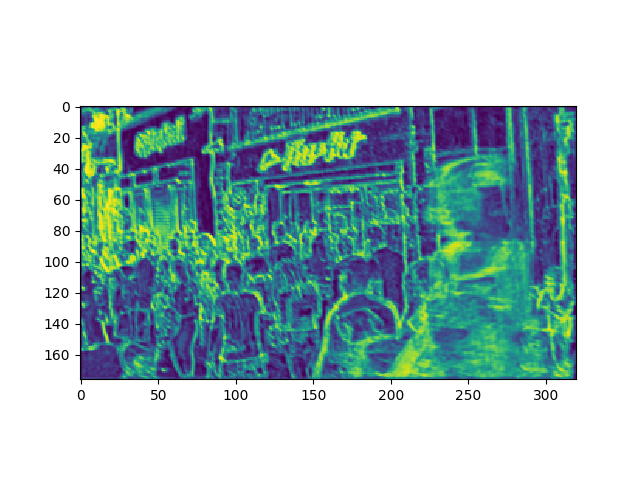}		
				\\ 				
				(a) Input Image&
				(b) Fusion $M_{fus}$& %\hspace{\fsdttwofig} &
				(c) Mask $M_1$  %\hspace{\fsdttwofig} &
				\\
	\end{tabular}
	%\vspace{-0em}
	\caption{Visualization of intermediate results on images from the GoPro test set~\cite{shen2019human}.} 
\label{fig:visualization1}
	%\vspace{-0em}
\end{figure}

\section{Experiments} 
\subsection{Implementation Details}
\noindent \textbf{Datasets:} We follow the configuration of~\cite{zhang2019deep,kupyn2019deblurgan,tao2018scale,kupyn2017deblurgan,nah2017deep}, which train on 2103 images from the GoPro dataset~\cite{nah2017deep}. For testing, we use two benchmarks: GoPro~\cite{nah2017deep} (1103 HD images), and HIDE \cite{shen2019human} (2025 HD images).

\noindent\textbf{Training settings and implementation details:} 
All the convolutional layers within our proposed modules contain $128$ filters. The hyper-parameters for our encoder-decoder backbone are $N=3$, $M=2$, and $P=2$, and filter size in PDF modules is $5\times5$. Following \cite{zhang2019deep}, we use batch-size of $6$ and patch-size of $256\times256$. Adam optimizer~\cite{kingma2014adam} was used with initial leaning rate $10^{-4}$, halved after every $2\times10^{5}$ iterations. We use PyTorch~\cite{paszke2017automatic} library and Titan Xp GPU.

\begin{table*}[htbp] \label{tbl:gopro}
\centering
\caption{Performance comparisons with existing algorithms on 1103 images from the deblurring benchmark GoPro \cite{nah2017deep}.\label{TableGopro}}
\resizebox{\textwidth}{!}{
\begin{tabular}{c|c|c|c|c|c|c|c|c|c|c|c|c|c}
\hline
Method & \cite{xu2013unnatural} & \cite{whyte2012non} & \cite{hyun2013dynamic} & \cite{gong2017motion} &  \cite{nah2017deep} & \cite{kupyn2017deblurgan} &  \cite{tao2018scale} &  \cite{zhang2018dynamic} & \cite{gao2019dynamic} & \cite{zhang2019deep} & \cite{kupyn2019deblurgan} & Ours(a) & Ours(b) \\
\hline
PSNR (dB) & 21 & 24.6 & 23.64 &  26.4  & 29.08 & 28.7 & 30.26 & 29.19 &30.90 & 31.20 &29.55 & \textcolor{red}{31.85} & \textbf{32.02} \\
SSIM & 0.741 & 0.846 & 0.824 & 0.863 & 0.914 & 0.858 & 0.934 & 0.931 &0.935 &0.940 &0.934 & \textcolor{red}{0.948} & \textbf{0.953}\\
Time (s) & 3800 & 700 & 3600 & 1200 & 6 & 1 & 1.2 &1 &1.0 &0.98 & 0.48 & \textbf{0.34} & \textcolor{red}{0.77} \\
%Size (MB) & - & - & - & 54.1 & 41.2 & 55 & 50 & 28 & 37.1 & 37.1 & 37.1 & 37.1 & \textbf{11.2} \\
\hline
\end{tabular}
}
\end{table*}

\begin{table}[htbp]
\centering
\caption{Performance comparisons with existing algorithms on 2025 images from the deblurring benchmark HIDE \cite{shen2019human}.\label{tbl:hide}}
\resizebox{\linewidth}{!}{
\begin{tabular}{c|c|c|c|c|c|c}
\hline
Method & \cite{kupyn2017deblurgan} & \cite{kupyn2019deblurgan} &  \cite{tao2018scale}  & \cite{shen2019human}\footnote{model trained on HIDE Dataset} & \cite{zhang2019deep} & Ours \\
\hline
PSNR & 24.51 & 26.61 & 28.36 & 28.89 & 29.09 & \textbf{29.98} \\
SSIM & 0.871 &0.875 & 0.915 & 0.930 & 0.924 & \textbf{0.930}\\
%Time (s) & 6 & 1 &1.0 &0.47 &0.35 &  \textbf{0.22} \\
%Size (MB) & - & - & - & 54.1 & 41.2 & 55 & 50 & 28 & 37.1 & 37.1 & 37.1 & 37.1 & \textbf{11.2} \\
\hline
\end{tabular}
}
\end{table}

\subsection{Performance comparisons}
The main application of our work is efficient deblurring of general dynamic scenes. Due to the complexity of the blur present in such images, conventional image formation model based deblurring approaches struggle to perform well. Hence, we compare with only two conventional methods \cite{whyte2012non,xu2013unnatural} (which are selected as representative traditional methods for non-uniform deblurring, with publicly available implementations). We provide extensive comparisons with state-of-the-art learning-based methods, namely MS-CNN\cite{nah2017deep}, DeblurGAN\cite{kupyn2017deblurgan}, DeblurGAN-v2\cite{kupyn2019deblurgan}, SRN\cite{tao2018scale}, and Stack(4)-DMPHN\cite{zhang2019deep}. We use official implementation from the authors with default parameters.

\textbf{Quantitative Evaluation}
We show performance comparisons on two different benchmark datasets. The quantitative results on GoPro testing set and HIDE Dataset \cite{shen2019human} are listed in Table 1 and 2. We evaluate two variants of our model with(b) and without(a) learnable offsets as shown in Table 1. %More results are included in our supplementary material. 

The average PSNR and SSIM measures obtained on the GoPro test split is provided in Table 1. It can be observed from the quantitative measures that our method performs better compared to previous state-of-the-art. The results shown in Figure 4. shows the large dynamic blur handling capability of our model while preserving sharpness.
% It achieves the best SSIM and PSNR on the test split.
We further evaluate the run-time of all the methods on a single GPU with images of resolution $720\times1280$. The  standard-deviation  of  the PSNR, SSIM, and run-time scores on the GoPro test set are 1.78, 0.018, and 0.0379, respectively. As reported in Table 1, our method takes significantly less time compared to other methods.

We also evaluate our method on the recent HIDE Dataset~\cite{shen2019human}. Both of GoPro and HIDE datasets contain dominant foreground object motion along with camera motion. We compare against all existing models trained on GoPro train-set for fair comparisons. As shown in Table 2, our approach outperforms all methods including \cite{shen2019human}, without requiring any human bounding box supervision. The superiority of our model is owed to the robustness of the proposed adaptive modules.

\textbf{Qualitative Evaluation:}
Visual comparisons on different dynamic and 3D scenes are shown in Figs.~\ref{fig:dynamic} and \ref{fig:visual_deblur_HIDE}.
Visual comparisons  are given in Fig.~\ref{fig:dynamic}. We observe that the results of prior works suffer from incomplete deblurring or artifacts. In contrast, our network is able to restore scene details more faithfully which are noticeable in the regions containing text, edges, etc. An additional advantage over \cite{hyun2013dynamic,whyte2012non} is that our model waives-off the requirement of parameter tuning during test phase.

% Benchmark Datasets The first row of Fig. 5 contains images from the testing datasets, which suffer from complex
% blur due to large camera and object motion.  Although traditional method [34] models a general non-uniform blur for
% camera  translation  and  rotation,  it  still  fails  for  Fig.  5(a),(c) and (d), where camera motion dominates. It is because
% forward/backward motion, as well as scene depth, plays important  roles  in  real  blurred  images.

On both the datasets, the proposed method achieves consistently better PSNR, SSIM and visual results with lower inference-time than DMPHN~\cite{zhang2019deep} and a comparable number of parameters.

%It is noteworthy that the existing deep deblurring methods~\cite{jongchan2018distort,hu2017exposure} do not perform well on this dataset because their advantage exists only for handling arbitrary local blur. For structured global blur, existing image formation model based methods perform well. 

\begin{table}[t]
\centering
\caption{Quantitative comparison of different ablations of our network on GoPro testset.
\label{TableAblationSingle}} 
\resizebox{0.45\textwidth}{!}{
\begin{tabular}{c|c|c|c|c|c|c}
\hline
Design &  $SA$&  $CA$ &  $CLA$ & $Kernel$ & $Offset$ & PSNR \\
\hline
Net1 & \xmark & \xmark & \xmark & \xmark & \xmark & 30.25\\
Net2 & \xmark & \xmark & \xmark & \checkmark & \xmark & 30.81\\
Net3 & \checkmark & \xmark & \xmark & \xmark & \xmark  & 30.76\\
Net4 & \checkmark & \checkmark & \xmark & \xmark  & \xmark  & 30.93\\
Net5 & \checkmark & \xmark & \checkmark & \xmark  & \xmark  & 31.12\\ %\hline
Net6 & \checkmark & \checkmark & \xmark & \checkmark & \xmark  & 31.44\\
Net7 & \checkmark & \checkmark & \checkmark & \checkmark & \xmark & 31.85\\ %\hline
Net8 & \checkmark & \checkmark & \checkmark & \checkmark & \checkmark  & 32.02 \\
\hline 
\end{tabular}
}
\label{tab:ablation}
\end{table}

\begin{figure*}[htb] \label{fig:visualization}
\begin{center}
\begin{tabular}{cccccc}
%\fbox{\rule{0pt}{2in} \rule{0.9\linewidth}{0pt}}
  \includegraphics[width=0.175\linewidth]{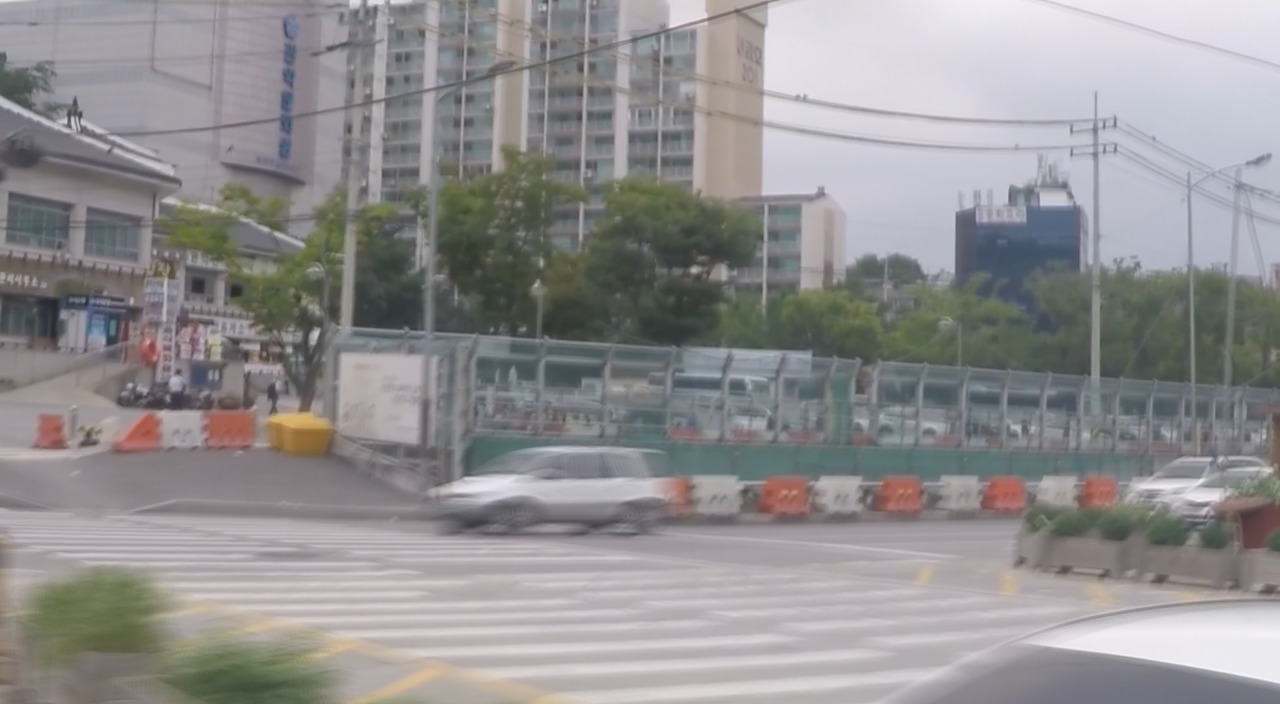} &
  \includegraphics[width=0.175\linewidth]{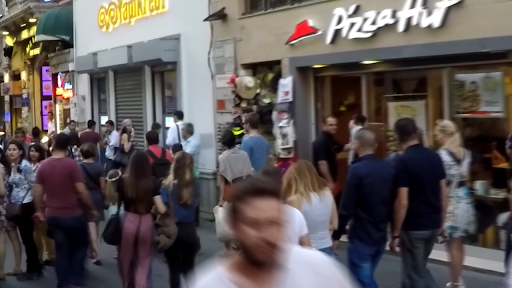} &
  \includegraphics[width=0.175\linewidth]{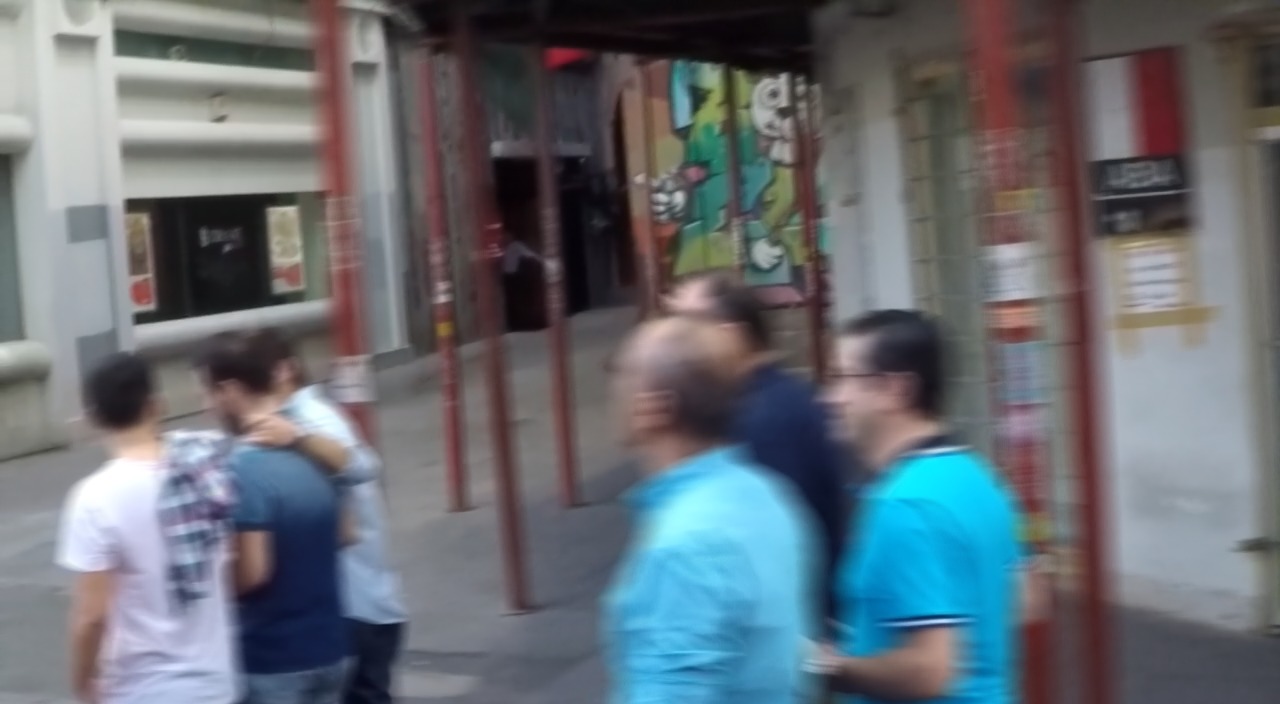} &
  \includegraphics[width=0.175\linewidth]{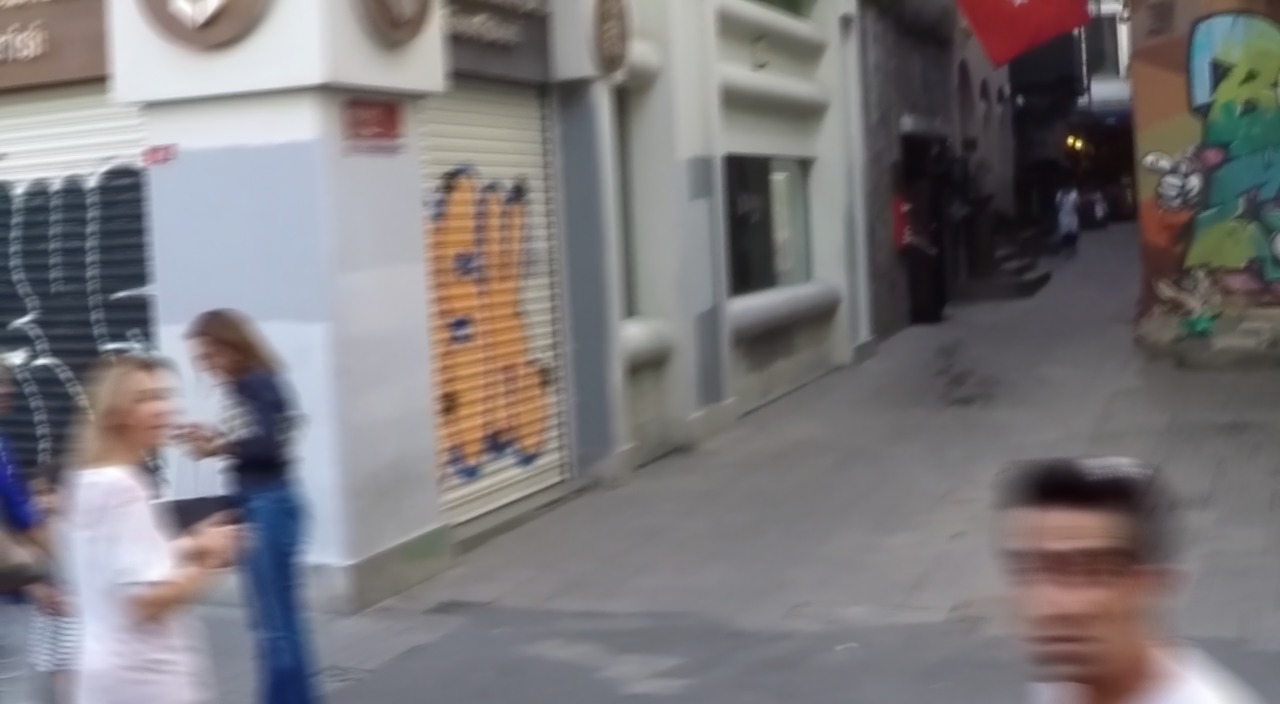} &
  \includegraphics[width=0.175\linewidth]{deformable_visualizations/004002} \\
  \includegraphics[width=0.175\linewidth]{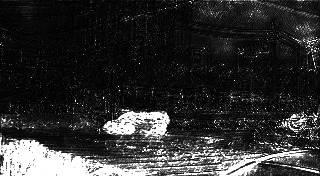}&
  \includegraphics[width=0.175\linewidth]{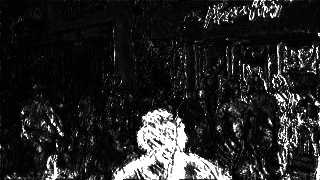} &
  \includegraphics[width=0.175\linewidth]{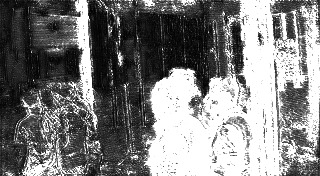} &
  \includegraphics[width=0.175\linewidth]{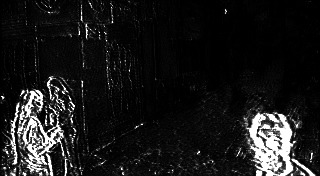} &
  \includegraphics[width=0.175\linewidth]{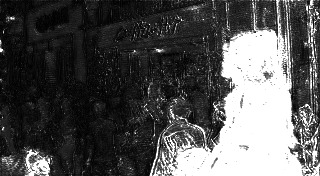} \\
  \includegraphics[bb=60 80 400 280,clip=True,width=0.175\linewidth]{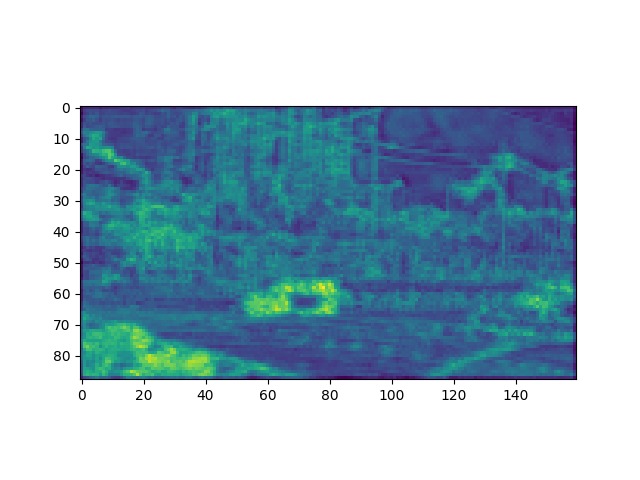}&
  \includegraphics[bb=60 80 400 280,clip=True,width=0.175\linewidth]{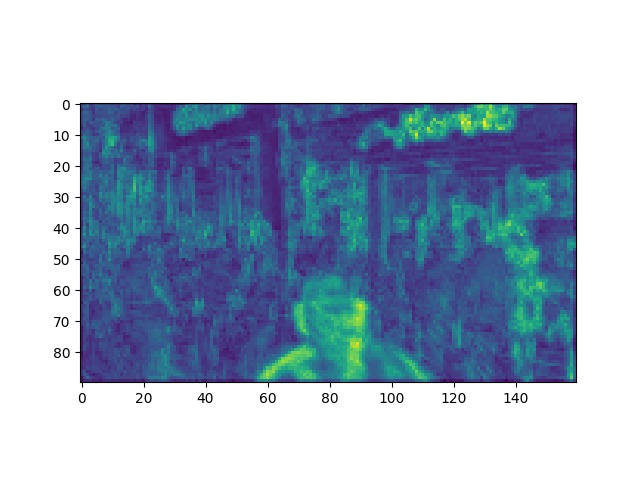}&
  \includegraphics[bb=60 80 400 280,clip=True,width=0.175\linewidth]{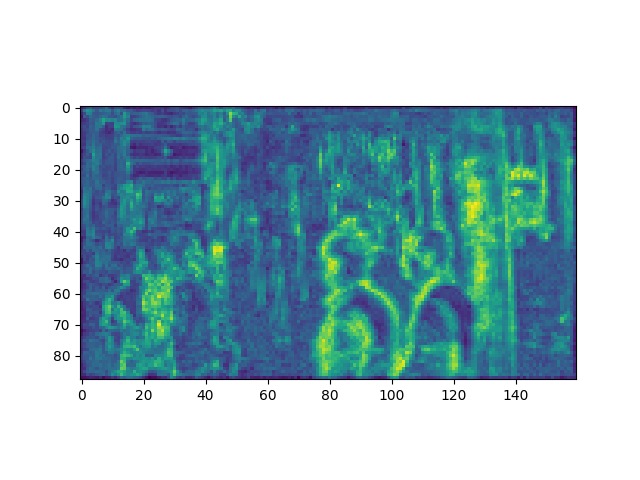}&
  \includegraphics[bb=60 80 400 280,clip=True,width=0.175\linewidth]{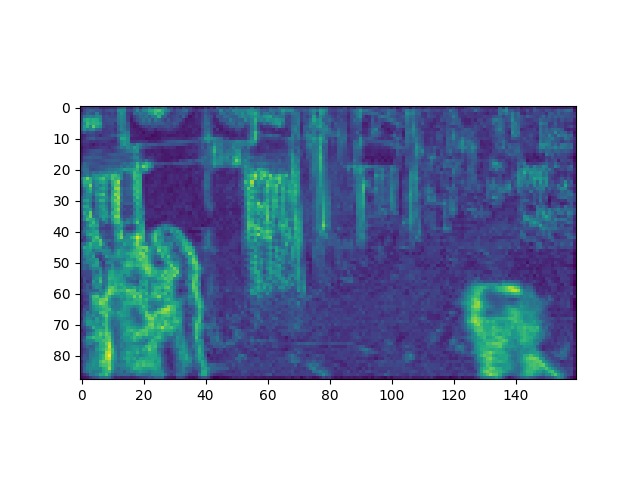}&
  \includegraphics[bb=60 80 400 280,clip=True,width=0.175\linewidth]{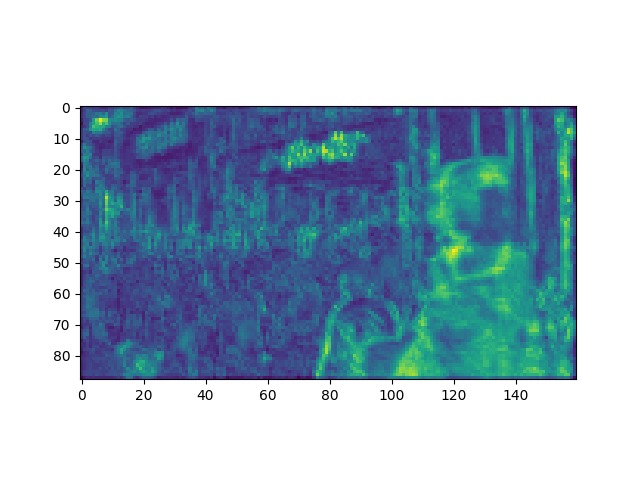}& \\
  \includegraphics[bb=60 80 400 280,clip=True,width=0.175\linewidth]{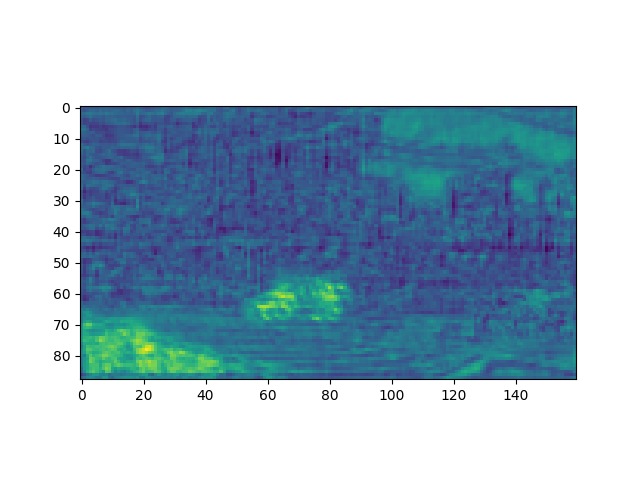}&
  \includegraphics[bb=60 80 400 280,clip=True,width=0.175\linewidth]{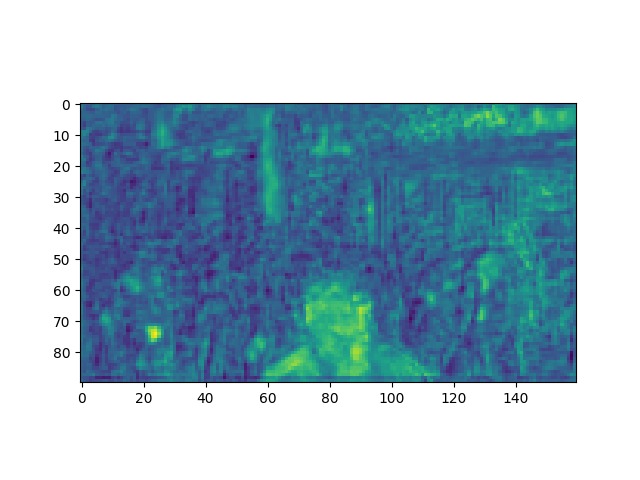}&
  \includegraphics[bb=60 80 400 280,clip=True,width=0.175\linewidth]{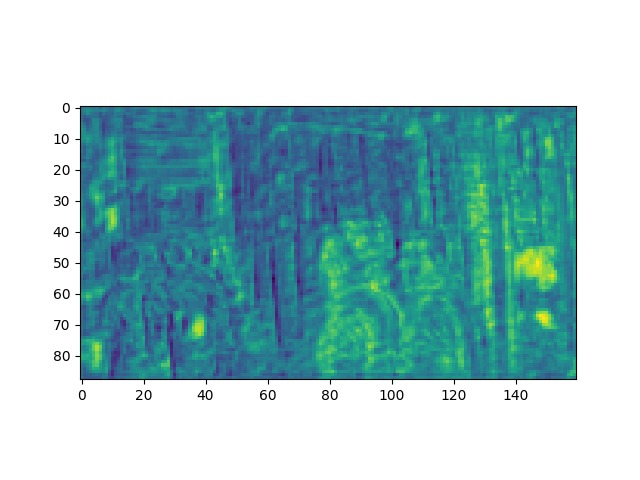}&
  \includegraphics[bb=60 80 400 280,clip=True,width=0.175\linewidth]{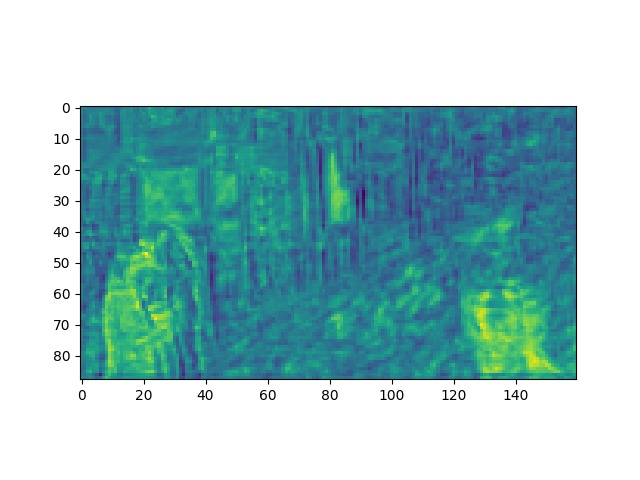}&
  \includegraphics[bb=60 80 400 280,clip=True,width=0.175\linewidth]{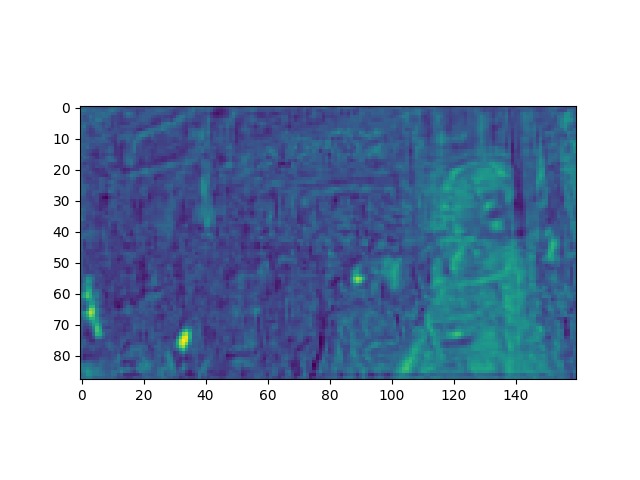}&
  \end{tabular}   \\
\end{center}
\vspace{-2.5mm}
\caption{The second row shows one of the spatial attention map for each image. The third row shows the spatial distribution of the horizontal-offset values for the filter. Fourth row shows the variance of the predicted kernel values.}
\label{fig:visualization}
\vspace{-4.5mm}
\end{figure*}
\subsection{Ablation studies}
% ESA efficient self attention module, CA cross attention module and CLA cross level attention module and PDF pixel-dependent filtering module 
In Table \ref{tab:ablation}, we analyse the effect of individual modules on our network's performance, using $1103$ test images from GoPro dataset~\cite{nah2017deep}. As shown in Figure \ref{fig:mainarch}, the proposed resblock contains one content-aware processing module and two standard convolutional layers. To find the optimal number of resblock in encoder and decoder we trained different versions of our network with varying number of resblocks. Although, the training performance as well as the quantitative results got better with the increase in number of blocks, beyond 3 the improvement was marginal. This led us to the choice of using 3 resblocks in each encoder and decoder and serves as a good balance between efficiency and performance as well.

As the use of local convolution and global attention together \cite{bello2019attention} or replacing local convolution with attention \cite{ramachandran2019stand} is explored recently for image recognition tasks, we further analyze it for image restoration tasks like deblurring. As shown in Table \ref{tab:ablation}, we observe that the advantages of SA and PDF modules are complimentary and their union leads to better performance (Net4 vs Net6). For better information flow between different layers of encoder-decoder and also between different levels we used CA, where the advantage of this attentive information flow rather than simple addition can be observed by comparing the performance of Net4 and Net5 compared to Net3. We also analyze the role of both adaptive weights and the adaptive local-neighborhood for PDF module. As shown quantitatively in Table \ref{tab:ablation} (Net7 and Net8) and visualized in Figure \ref{fig:visualization}, adaptiveness of the offsets along with the weights perform better as it satisfies the need of directional local filters. We have also showed comparisons of the convergence plots of these models in supplementary. We also try to incorporate the attention mechanism used in \cite{bello2019attention} in our model for fair comparison. Due to high memory requirement, we were only able to use one attention module in the decoder in each level. The resultant PSNR was 30.52 compared to 30.76 of Net3. But, as it already occupied full GPU memory, we were unable to introduce more blocks, or cross attention.
% Note that although the proposed network is already quite efficient, replacing the standard convolutions in our network with grouped convolution and/or separable convolution can lead to further reduction in model size and inference time. We leave this analysis for future work.

\subsection{Visualization and Analysis}
The first row of Fig. \ref{fig:visualization} contains images from the testing datasets which suffer from complex blur due to large camera and object motion. In the subsequent rows, we visualize the output of different modules of our network and analyze the behavior change while handling different levels of blur due to camera motion, varying depth, moving objects, etc. The second row of Fig. \ref{fig:visualization} shows one of the attention-maps ($q_i$, $i \in {1,2,...C_2}$) corresponding to each image. We can observe the high correlation between estimated attention weights and the dominant motion blurred regions present in the image. This adaptive ability of the network to focus on relevant parts of the image can be considered crucial to the observed performance improvement. The third and fourth rows of Fig. \ref{fig:visualization} show the spatially-varying nature of filter weights and offsets. Observe that a large horizontal offset is estimated in the regions with high horizontal blur so that the filter shape can spread along the direction of motion. Although the estimated filter wights are not directly interpretable, it can be seen that the variance of the filters correlates with the magnitude of blur. We further visualize the behavior of the fusion mask which adaptively weighs the outputs of the two branches for each pixel location. As shown in Fig. 6, PDF module output is more preferred in regions with moving foreground objects or blurred edges where most of the other regions give almost equal weight to both the branches. On the other hand, homogeneous regions where the effect of blur is negligible, have shown a preference towards the attention branch. To further investigate this behavior, we have visualized the spatial mask ($M_1$). As we can observe in Fig. 6(c), the mask suppresses these homogeneous regions even before calculating self-attention for each pixel. This shows the robustness and interpretability of our attention module while handling any type of blur.

\noindent \textbf{PDF Module:} We synthetically blurred 25 sharp images using synthetic linear PSFs oriented in 4 different directions (0$\degree$,45$\degree$,90$\degree$,135$\degree$). For these images, we recorded the dominant direction of filter offsets estimated by our PDF module. The values obtained (11$\degree$,50$\degree$,81$\degree$,126$\degree$) show high correlation between the offset orientations and the PSF angles.

\section{Conclusions}
We proposed a new content-adaptive architecture design for the challenging task of removing spatially-varying blur in images of dynamic scenes. Efficient self-attention is utilized in all the encoder-decoder to get better representation whereas cross-attention helps in efficient feature propagation across layers and levels. Proposed dynamic filtering module shows content-awareness for local filtering. The complimentary behaviour of the two branches are shown in Table \ref{TableAblationSingle} and Fig. 6. Different from existing deep learning-based methods for such applications, the proposed method is more interpretable which is one of its key strengths. Our experimental results demonstrated that the proposed method achieved better results than state-of-the-art methods on two benchmarks both qualitatively and quantitatively. We showed that the proposed content-adaptive approach achieves an optimal balance of memory, time and accuracy and can be applied to other image-processing tasks. 
% The interpretability of deep learning has been attracting much attention~\cite{selvaraju2017grad}, and it is especially important for some applications such as medical image restoration~\cite{razzak2018deep}.

{\small
\bibliographystyle{ieee_fullname}
\bibliography{egbib}
}

\end{document}